\documentclass[lettersize,journal, subfigure]{IEEEtran}
\usepackage{amsmath,amsfonts}
\usepackage{array}
\usepackage{subfigure}
\usepackage{textcomp}
\usepackage{stfloats}
\usepackage{url}
\usepackage{verbatim}
\usepackage{graphicx}
\usepackage{multicol}
\usepackage{multirow}
\usepackage{amssymb}
\usepackage{booktabs}
\usepackage{indentfirst}                                                                
\usepackage{bbding}
\usepackage{makecell}
\usepackage[dvipsnames, table]{xcolor}
\usepackage{color}

\usepackage{diagbox}
\usepackage{arydshln}
\usepackage{hyperref}
\usepackage{cleveref}
\usepackage[ruled, lined, linesnumbered, commentsnumbered, longend]{algorithm2e}
\usepackage{amsmath}
\usepackage{tabularx}
\usepackage{amssymb}

\usepackage{float}

\DeclareMathOperator*{\argmax}{arg\,max}
\DeclareMathOperator*{\argmin}{arg\,min}

\newcommand{\tab}[1]{{\color{Purple} #1}}
\newcommand{\revised}[1]{{\color{black} #1}}
\newcommand{\xgy}[1]{{\color{black} #1}}

\newcommand{\ljq}[1]{{\color{black} #1}}
\def\BibTeX{{\rm B\kern-.05em{\sc i\kern-.025em b}\kern-.08em
    T\kern-.1667em\lower.7ex\hbox{E}\kern-.125emX}}
\usepackage{balance}

\begin{document}
\title{IM-IAD: Industrial Image Anomaly Detection Benchmark in Manufacturing}
\author{Guoyang Xie$^{1}$,~\IEEEmembership{Member,~IEEE,}
        Jinbao Wang$^{1}$,~\IEEEmembership{Member,~IEEE,}
        Jiaqi Liu$^{1}$,
        Jiayi Lyu,
        Yong Liu,
        Chengjie Wang, 
        Feng Zheng$^{*}$,~\IEEEmembership{Member,~IEEE,} and
        Yaochu Jin$^{*}$,~\IEEEmembership{Fellow,~IEEE}
\thanks{Guoyang Xie is with the Department of Computer Science and Engineering, Southern University of Science and Technology, Shenzhen 518055, China and is also with the Department of Computer Science, University of Surrey, Guildford GU2 7YX, United Kingdom (e-mail: guoyang.xie@ieee.org)}
\thanks{Jinbao Wang, Jiaqi Liu and Feng Zheng are with the Department of Computer Science and Engineering, Southern University of Science and Technology, Shenzhen 518055, China (e-mail: wangjb@ieee.org; liujq32021@mail.sustech.edu.cn; f.zheng@ieee.org)}
\thanks{Yong Liu and Chengjie Wang are with Tencent Youtu Lab, Shenzhen 518040, China (e-mail: chaosliu@tencent.com; jasoncjwang@tencent.com)}
\thanks{Jiayi Lyu is with the School of Engineering Science, University of Chinese Academy of Sciences, Beijing, China (e-mail: lyujiayi21@mails.ucas.ac.cn)}
\thanks{Yaochu Jin is with the School of Engineering, Westlake University, Hangzhou 310030, China and also with the Department of Computer Science and Engineering, University of Surrey, Guildford GU2 7YX, United Kingdom (e-mail: jinyaochu@westlake.edu.cn )}
\thanks{$^1$Contributed equally.}
\thanks{$*$Co-corresponding authors.}
}

\markboth{Journal of \LaTeX\ Class Files,~Vol.~18, No.~9, September~2020}%
{How to Use the IEEEtran \LaTeX \ Templates}

\maketitle

\begin{abstract}

 \xgy{Image anomaly detection (IAD) is an emerging and vital computer vision task in industrial manufacturing (IM). Recently, many advanced algorithms have been reported, but their performance deviates considerably with various IM settings. We realize that the lack of a uniform IM benchmark is hindering the development and usage of IAD methods in real-world applications. In addition, it is difficult for researchers to analyze IAD algorithms without a uniform benchmark. To solve this problem, we propose a uniform IM benchmark, for the first time, to assess how well these algorithms perform, which includes various levels of supervision (unsupervised versus fully supervised), learning paradigms (few-shot, continual and noisy label), and efficiency (memory usage and inference speed). Then, we construct a comprehensive image anomaly detection benchmark (IM-IAD), which includes 19 algorithms on seven major datasets with a uniform setting. Extensive experiments (17,017 total) on IM-IAD provide in-depth insights into IAD algorithm redesign or selection. Moreover, the proposed IM-IAD benchmark challenges existing algorithms and suggests future research directions. For reproducibility and accessibility, the source code is uploaded to the website: \href{https://github.com/M-3LAB/open-iad}{https://github.com/M-3LAB/open-iad} }

\end{abstract}

\begin{table}[ht]
\centering
\caption{The comparison with IAD and related benchmarks in terms of dataset, learning paradigm, algorithm and metric.}
\large
\renewcommand{\arraystretch}{1.3}
\resizebox{0.48\textwidth}{!}{
\begin{tabular}{l|l|l|c|c|c}
\hline
\rowcolor{blue!10} \textbf{Category}         &  \multicolumn{2}{c|}{\textbf{Classification / Attribute}}    &       \multicolumn{3}{c}{\textbf{Mark}}    \\
\hline
 Work         &        \multicolumn{2}{l|}{--}      &  \cite{Zheng2022BenchmarkingUA} &  \cite{Han2022ADBenchAD} & IM-IAD \\
\hline
\multirow{8}{*}{Dataset}        &  \multicolumn{2}{l|}{MVTec AD~\cite{Bergmann2019MVTecA, bergmann2021mvtecIJCV}}                &  \textcolor{blue}{\checkmark}       &   \textcolor{blue}{\checkmark}  &  \textcolor{blue}{\checkmark}      \\ \cdashline{2-6}[1pt/1pt]
                                &  \multicolumn{2}{l|}{MVTec LOCO-AD~\cite{bergmann2022beyond}}            &   $\backslash$    &   $\backslash$  &  \textcolor{blue}{\checkmark}        \\ \cdashline{2-6}[1pt/1pt]
                                &  \multicolumn{2}{l|}{BTAD~\cite{mishra2021vt}}                     &  $\backslash$     &  $\backslash$   &  \textcolor{blue}{\checkmark}        \\ \cdashline{2-6}[1pt/1pt]
                                &  \multicolumn{2}{l|}{MPDD~\cite{jezek2021deep}}                    &    $\backslash$   & $\backslash$    &  \textcolor{blue}{\checkmark}        \\ \cdashline{2-6}[1pt/1pt]
                                &  \multicolumn{2}{l|}{MTD~\cite{huang2020surface}}                      &     $\backslash$  &   $\backslash$  &  \textcolor{blue}{\checkmark}        \\ \cdashline{2-6}[1pt/1pt]
                                &  \multicolumn{2}{l|}{VisA~\cite{zou2022spot}}                     &    $\backslash$   &  $\backslash$   &  \textcolor{blue}{\checkmark}        \\ \cdashline{2-6}[1pt/1pt]
                                &  \multicolumn{2}{l|}{DAGM~\cite{DAGMGNSS2077}}                     &    $\backslash$   &   $\backslash$  &  \textcolor{blue}{\checkmark}        \\
\hline
\multirow{5}{*}{Paradigm}       & \multicolumn{2}{l|}{Unsupervised}         &  \textcolor{blue}{\checkmark}      &  \textcolor{blue}{\checkmark}     &  \textcolor{blue}{\checkmark}        \\ \cdashline{2-6}[1pt/1pt]
                                 & \multicolumn{2}{l|}{Fully Supervised}      &  $\backslash$     &  $\backslash$     &  \textcolor{blue}{\checkmark}        \\\cdashline{2-6}[1pt/1pt]
                                & \multicolumn{2}{l|}{Few-Shot}          &    $\backslash$   &   $\backslash$  &  \textcolor{blue}{\checkmark}        \\ \cdashline{2-6}[1pt/1pt]
                                & \multicolumn{2}{l|}{Noisy Label}               &  $\backslash$     &  \textcolor{blue}{\checkmark}   &  \textcolor{blue}{\checkmark}        \\ \cdashline{2-6}[1pt/1pt]
                                & \multicolumn{2}{l|}{Continual}              &    $\backslash$   &   $\backslash$    &  \textcolor{blue}{\checkmark}        \\ 
\hline
\multirow{6}{*}{Algorithm} & \multirow{4}{*}{\makecell[l]{Feature\\ Embedding}} & Normalizing Flow                    &  \textcolor{blue}{\checkmark}       &   $\backslash$  &  \textcolor{blue}{\checkmark}        \\ \cdashline{3-6}[1pt/1pt]
                                &                                    & Memory Bank              &  \textcolor{blue}{\checkmark}       &   $\backslash$  &  \textcolor{blue}{\checkmark}        \\ \cdashline{3-6}[1pt/1pt]
                                &                                    & Teacher-Student          &  \textcolor{blue}{\checkmark}       &  $\backslash$   &  \textcolor{blue}{\checkmark}        \\ \cdashline{3-6}[1pt/1pt]
                                &                                    & One-Class Classification         &  \textcolor{blue}{\checkmark}       &  $\backslash$   &  \textcolor{blue}{\checkmark}        \\ \cdashline{2-6}[1pt/1pt]
                                & \multirow{2}{*}{Reconstruction}    & External Data            &  \textcolor{blue}{\checkmark}       & $\backslash$    &  \textcolor{blue}{\checkmark}        \\ \cdashline{3-6}[1pt/1pt]
                                &                                    & Internal Data            &  \textcolor{blue}{\checkmark}       &  $\backslash$   &  \textcolor{blue}{\checkmark}        \\ \cdashline{3-6}[1pt/1pt]
\hline
\multirow{7}{*}{Metric}         & AUROC             & Image- and Pixel-Level              &  \textcolor{blue}{\checkmark}       &  $\backslash$   &  \textcolor{blue}{\checkmark}        \\ \cdashline{2-6}[1pt/1pt]
                                & AUPR/AP               & Image- and Pixel-Level              &  \textcolor{blue}{\checkmark}       &   $\backslash$  &  \textcolor{blue}{\checkmark}        \\ \cdashline{2-6}[1pt/1pt]
                                & PRO                                &          Pixel-Level                &    $\backslash$   &   $\backslash$  &  \textcolor{blue}{\checkmark}        \\ \cdashline{2-6}[1pt/1pt]
                                & SPRO                               &               Pixel-Level           &   $\backslash$    &  $\backslash$   &  \textcolor{blue}{\checkmark}        \\ \cdashline{2-6}[1pt/1pt]
                                & FM                                 &               Image- and Pixel-Level           &    $\backslash$   &  $\backslash$   &  \textcolor{blue}{\checkmark} 
                                \\ \cdashline{2-6}[1pt/1pt]
                                & Efficiency                         & Inference Speed           &  \textcolor{blue}{\checkmark}       & $\backslash$    &  \textcolor{blue}{\checkmark} \\ \hline
                                
 \rowcolor{black!10}\textbf{Uniform}     & \multicolumn{2}{l|}{--}   &   $\backslash$       &  $\backslash$   &  \textcolor{blue}{\checkmark}        \\ \hline
\end{tabular}
}
\label{tab:benchmark_comparison_table}
\end{table}

\section{Introduction}\label{sec:intro}

 IAD is an important computer vision task for industrial manufacturing (IM) applications~\cite{Rathore2020VisualSA,wang2021dynamic, Wang2022HybridVM}, such as industrial products surface anomaly detection~\cite{Yoo2019ConvolutionalRR,huang2020surface}, textile defect detection~\cite{tsang2016fabric,li2021fabric}, and food inspection~\cite{Bonfiglioli2022TheED,zhang2023pku}. \xgy{However, few IAD algorithms are used in real industrial manufacturing and there is an urgent demand for a uniform benchmark for image anormaly detection (IAD) to bridge the gap between adacademic research and practical applications. Despite this, current research in the computer vision community primarily focuses on unsupervised learning, and little effort has been dedicated to the analysis of the industry's demands. For the above reason, it is important and urgent to build a uniform benchmark for IM}.

\xgy{The proposed IM-IAD aims to push the boundaries of IAD methods in practical scenarios, since existing benchmarks cannot accurately reflect the needs of IM.} There are two notable pieces of work~\cite{Zheng2022BenchmarkingUA, Han2022ADBenchAD} that make efforts to benchmark IAD algorithms. The distinctions between IM-IAD and existing benchmarks lie in the following aspects.
Firstly, previous studies mainly concentrate on benchmarking IAD methods based on the level of supervision~\cite{Han2022ADBenchAD}. \xgy{However, these work~\cite{Han2022ADBenchAD,Zheng2022BenchmarkingUA} ignore realistic IM requirements, e.g., continual IAD, few-shot IAD, and noisy label IAD and the trade-off between accuracy and efficiency. It is crucial because most existing IAD algorithms cannot meet the above-mentioned requirements of IM. There are several reasons. Firstly, IAD algorithms prioritize a higher accuracy but ignore the inference speed and GPU memory size, which hinders IAD algorithms from being used in factories. Secondly, it is tough to collect many normal samples for training due to commercial privacy; the performance of IAD algorithms hardly meets the requirements of IM if the number of training samples is limited. Thirdly, most existing IAD algorithms suffer from catastrophic forgetting as they do not possess continual learning abilities. Finally, wrong annotations inevitably happen due to the tiny size of the defect. As a result, some abnormal samples in the training dataset are labelled as normal. The nosiy training dataset will result in degradation of the algorithms' performance. But most existing IAD algorithms do not consider the noisy labeling issue. Hence, our purposed IM-IAD makes researchers aware of the gap between academia and industry and offers deeper insights into future improvements. Secondly, ADBench~\cite{Han2022ADBenchAD} focuses primarily on tabular and graph-structured data instead of image data and no IAD algorithms have been evaluated on the benchmark. For instance in the most recent work\cite{Zheng2022BenchmarkingUA}, the authors only assess unsupervised IAD algorithms on two datasets. By contrast, we constructed a comprehensive benchmark, IM-IAD, with seven industrial datasets and 19 algorithms, as shown in Table~\ref{tab:IAD_alg}. }


In this paper, we address the above issues in IAD through extensive experiments. The \textit{\textbf{key takeaways}} are as follows. 1) Regarding accuracy, memory usage, and inference speed, none of the benchmarked unsupervised IAD algorithms is statistically significantly better than others, highlighting the importance of selecting the types of anomaly. 2) The long-distance attention mechanism shows great potential in logical IAD, possibly due to its global feature extraction abilities. 3) Fully supervised methods have demonstrated superior performance compared to unsupervised IAD. This can be attributed to the fact that the incorporation of labeled anomalies into the training process significantly enhances IAD capabilities. 4) With merely 4 augmented (rotated) data, feature embedding-based few-shot IAD algorithms can achieve the performance 95\% of vanilla IAD, revealing the necessity of data characteristics. 5) Importance reweighting successfully improves the resilience of IAD algorithms, even though the noise ratio is larger than 10\%. 6) Memory bank can be seamlessly incorporated into advanced IAD algorithms, considerably enhancing their capacity to resist catastrophic forgetting.

The main contributions are summarized as follows.
\begin{itemize}
    \item We extract scientific problems from the manufacturing process and present a standardized and uniform benchmark to bridge the gap between academic research and industrial practices in the identification of image anomalies.
    \item We examine 16 IAD methods on 7 benchmark datasets, resulting in a total of 17,017 instances. Additionally, we present a plug-and-play and modular implementation for fair IAD evaluation, which greatly benefits the future development of IAD algorithms.
    \item By analyzing the requirements of research and industrial manufacturing processes, we examine four key aspects of IAD algorithms for comparison: the changeover-based few-shot representational abilities; the trade-off between accuracy and efficiency; the catastrophic forgetting phenomenon; and the robustness of the algorithm in the presence of noise labeling. Based on these aspects, we offer deep insights and suggest future directions.
\end{itemize} 

The rest of this paper is organized as follows. Sec.~\ref{sec:related-work} presents a literature review on IAD methods and mainstream learning paradigms. Sec.~\ref{sec:IMIAD-method} provides our proposed uniform benchmark, IM-IAD. In detail, we give comprehensive experimental evaluations in Sec.~\ref{sec:benchmark_result} and discuss the advanced model performance in various IM scenarios. Finally, Sec.~\ref{sec:conclusion} draws the conclusion of our work.

\begin{figure*}[th]
    \centering
    \includegraphics[width=1\linewidth]{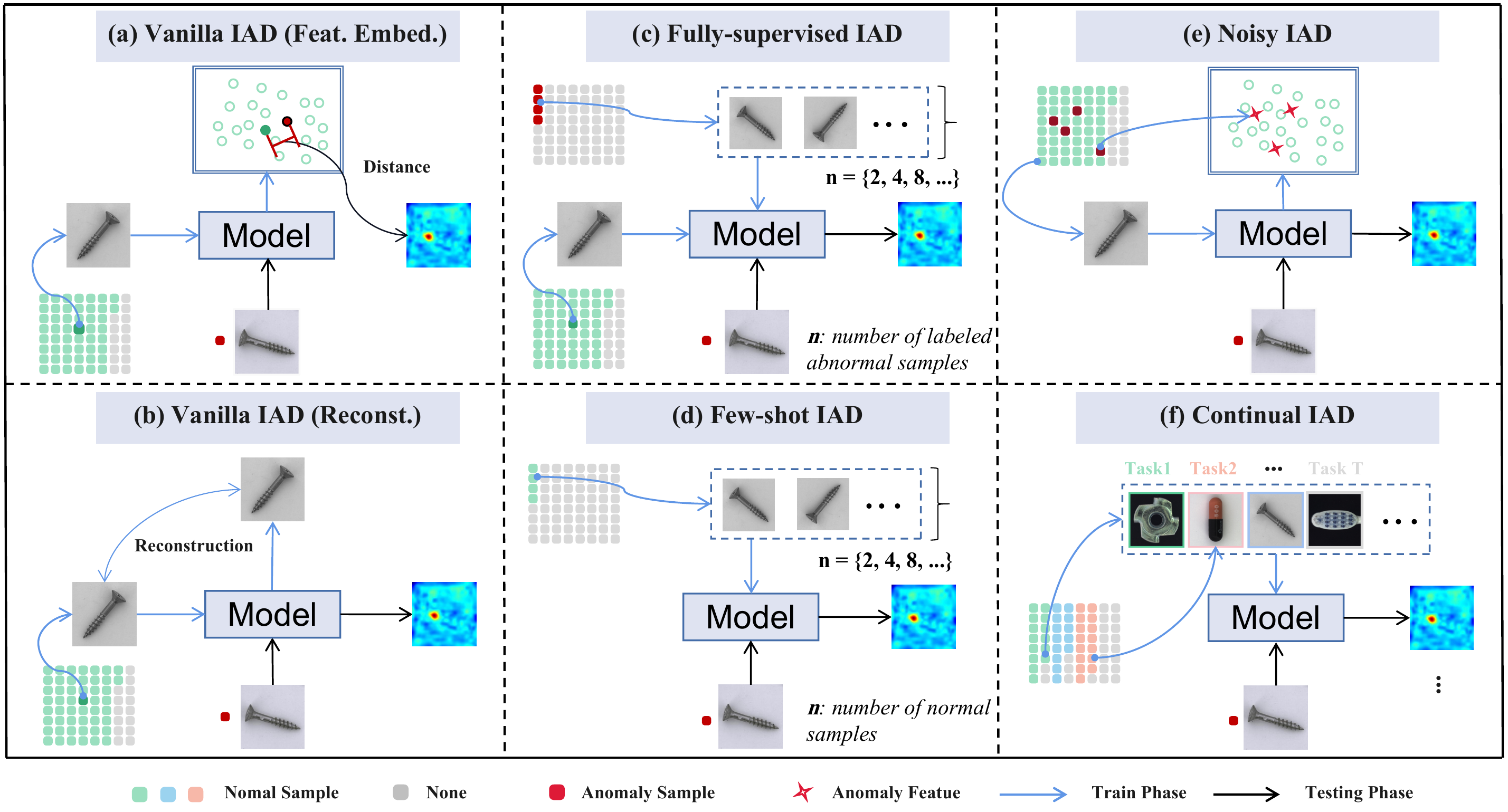}
    \caption{Illustration of the IM-IAD. The vanilla unsupervised IAD methods can be divided into two categories, namely feature embedding-based and reconstruction-based methods. (a) Feature embedding-based methods find the difference between the test samples and normal samples at the feature level, while (b) reconstruction-based methods compare the difference between the input image and the reconstructed image to determine whether it is abnormal or not. For fully supervised methods (c), they use limited abnormal samples with annotations to improve the model performance. The few-shot setting (d) uses a limited number of normal samples for training. The noisy setting (e) mixes abnormal samples in the training set and evaluates the robustness of the model. The continual setting (f) trains on each task in turn and evaluates how much the model forgets past tasks. 
    }
    \label{fig:fewshot_ad_paradigm}
\end{figure*}

\section{Related Work} \label{sec:related-work}

This section has reviewed the current IAD methods concluded in Table~\ref{tab:benchmark_comparison_table} with five settings that are visualized in Fig.~\ref{fig:fewshot_ad_paradigm} and are defined in Sec.~\ref{sec:definition}.

\textbf{Unsupervised IAD.} \xgy{There are two streams methods on unsupervised IAD~\cite{liu2023deep}, namely feature embedding-based methods~\cite{bergmann2020uninformed,rudolph2021same,cohen2020sub,cao2024bias}, and reconstruction-based methods~\cite{yang2020dfr,zavrtanik2021draem}}\ljq{,~\cite{zhang2023unsupervised}}. Specifically, feature embedding-based approaches can be divided into four categories, including teacher-student~\cite{bergmann2020uninformed}, normalizing flow~\cite{rudolph2021same}, memory bank~\cite{cohen2020sub}, and one-class classification~\cite{sohn2020learning}. The most typical methods are teacher-student models and memory bank-based models. As for teacher-student models,  the teacher model extracts the features of normal samples and distils the knowledge to the student model during the training phase. Regarding abnormal images, the features extracted by the teacher network may deviate from those of the student network during the test phase. Thus, the basic rule for finding anomalies is that the teacher-student network has different features. Regarding memory bank-based approaches~\cite{cohen2020sub,li2021anomaly,kim2021semi}, the models capture the features of normal images and store them in a feature memory bank. During the testing phase, the feature of the test sample queries the memory bank for the feature points of k-nearest neighbourhoods. The test sample is abnormal if the distance between the test feature and the closest feature points of neighbourhoods exceeds a specific threshold. \ljq{However, both of them heavily depend on the teacher network's power or the memory bank's size, which may limit the generalization ability in the real-world industry.} \ljq{Reconstruction-based methods, from those using Autoencoder~\cite{bergmann2018improving,zavrtanik2021draem,chen2023easynet} to Generative Adversarial Network~\cite{yan2021learning,duan2023few}, to Transformer~\cite{you2022unified,yao2023focus} and the diffusion model~\cite{lu2023removing,zhang2023unsupervised}, have been employed in recent years. These methods require a significant amount of training time. However, their performance still lags behind feature embedding-based methods, making it challenging to meet the demands of practical industrial production.}

\textbf{Fully supervised IAD.} Regarding the setting, the distinction between unsupervised IAD and fully supervised IAD~\cite{li2023efficient,chu2020neural, ding2022catching} is the use of abnormal images for training. Fully supervised IAD methods~\cite{venkataramanan2020attention,pang2021explainable} focus on efficiently employing a small number of anomalous samples to distinguish the features of abnormalities from those of normal samples. Nevertheless, the performance of some fully supervised IAD approaches is inferior to that of unsupervised methods for identifying anomalies. \ljq{There is still much room for improvement regarding the use of data from abnormal samples.} 

\revised{
\textbf{Few-shot IAD.} 
Few-shot (including zero-shot) IAD is very promising because it can significantly reduce the demand for data volume. However, it is still in its infancy. GraphCore~\cite{xie2023pushing} designs efficient visual isometric invariant features for a few-shot task, which can perform fast training and significantly improve the ability to discriminate anomalies using a few samples. \ljq{FastRecon~\cite{fang2023fastrecon} stores the features of normal samples to assist image reconstruction during the inference process, enabling the training of the reconstruction network in a few-shot learning scenario. WinCLIP~\cite{jeong2023winclip} extends the feature comparison approach to zero-shot scenarios. It uses CLIP~\cite{radford2021learning} to extract text features of normal or abnormal descriptions. It compares them with the test image features for anomaly detection.} Furthermore, by leveraging diverse multi-modal prior knowledge of foundation models (like Segment Anything~\cite{kirillov2023segment}) for anomalies, \ljq{Cao~\textit{et al.}~\cite{cao2023segment} achieve comparable performance on several benchmarks in the zero-shot setting. Recently, AnoVL~\cite{deng2023anovl} and AnomalyGPT~\cite{gu2023anomalygpt} have achieved new state-of-the-art results in zero-shot scenarios using the understanding capabilities of visual language models, but the size of these models is too large to be deployed in real production lines.} Due to the real needs of industrial scenarios, few-shot learning remains a future research focus in anomaly detection.

\textbf{Noisy IAD.} 
Noisy learning is a classic problem for anomaly detection~\cite{qiu2022latent, Chen2022DeepOC}. However, in the field of IAD, training with noisy data is an inevitable problem in practice. Such noise usually comes from inherent data shifts or human misjudgments. To address this problem, SoftPatch~\cite{jiang2022softpatch} is the first to efficiently denoise data at the patch level in an unsupervised manner. \ljq{InReaCh~\cite{mcintosh2023inter} strengthens the model's representation ability by learning the internal relationships within the image. It can achieve results close to SOTA even when trained on artificially corrupted data.} By contrast, other methods rarely discuss this situation. The IAD with noisy data needs more research in the future.
}

\textbf{Continual IAD.} Work integrating continual learning (CL) is growing with the development of anomaly detection. The first CL benchmark for industrial IAD is introduced by Li~\textit{et al.}~\cite{li2022towards}, but their setting ignores the large domain gap of different datasets. \ljq{LeMO~\cite{gao2023towards} focuses on the problem of continual learning within a class, which, however, neglects the issue of forgetting between classes.} To further explore the ability of IAD methods, more efforts should be made in this setting.

\section{IM-IAD}\label{sec:IMIAD-method}

In this section, we first present the definition of five settings in IM-IAD and then summarize the implementation details of baseline methods, mainstream datasets, evaluation metrics, and hyperparameters. Finally, we point out the importance of studying IM-IAD from a uniform perspective.

\subsection{Problem Definition} \label{sec:definition}
The goal of IAD is that, given a normal or abnormal sample from a target category, the anomaly detection model should predict whether or not the image is anomalous and localize the anomaly region if the prediction result is abnormal. 
We provide the following five settings.
\begin{enumerate}
\item \textbf{Unsupervised IAD.} The training set consists only of $m$ normal samples for each category. The test set contains normal and abnormal samples.

\item \textbf{Fully supervised IAD.} The training set consists of $m$ normal samples and $n$ abnormal samples, where $n \ll m$. In our case, we set $n$ as 10.
    
\item \textbf{Few-shot IAD.} Given a training set of only $m$ normal samples, where $m \leq  8$, from a certain category. The number of $m$ could be 1, 2, 4, and 8 for the target category, respectively.
    
\item \textbf{Noisy IAD.} Given a training set of $m$ normal samples and $n$ abnormal samples, $n$ most of 20\% of $m+n$. And $n$ abnormal samples are labeled as normal samples in the training dataset.
    
\item \textbf{Continual IAD.} Given a finite sequence of training dataset $\mathcal{T}_{train}^{total}$ consists of $n$ categories, $\mathcal{T}_{train}^{total} = \left\{\mathcal{T}_{train}^{1}, \mathcal{T}_{train}^{2}, \cdots, \mathcal{T}_{train}^{n} \right\}$, \textit{i.e.}, $\mathcal{T}_{train}^{total} = \bigcup^{n}_{i=1} \mathcal{T}_{train}^{i}$, where the subsets $\mathcal{T}_{train}^{i}$ consists of normal samples from one certain category $c_{i}, i\in n$. The IAD algorithm is trained once for each category dataset $\mathcal{T}_{train}^{i}$ in the CL scenario. During the test, the updated model is evaluated on each category of previous datasets $\mathcal{T}_{test}^{total}$, \textit{i.e.}, $\mathcal{T}_{test}^{total} = \left\{\mathcal{T}_{test}^{1}, \mathcal{T}_{test}^{2}, \cdots,  \mathcal{T}_{test}^{i-1} \right\}$, respectively. 
\end{enumerate}

\begin{table*}[t]
\centering
\caption{Representative algorithms for IM-IAD. The purple ones indicate our re-implementation methods.}
\renewcommand{\arraystretch}{1.3}
\resizebox{1\linewidth}{!}{
\begin{tabular}{l|l|l|l}
\hline
\rowcolor{blue!5} \multicolumn{3}{l|}{\textbf{Paradigm}} & \textbf{Methods} \\
\hline
  \multirow{7}{*}{\textbf{Vanilla}} & \multirow{5}{*}{Feature embeding} & Normalizing flow    &  \tab{CS-Flow}~\cite{rudolph2022fully}, \tab{FastFlow}~\cite{yu2021fastflow}, CFlow~\cite{gudovskiy2022cflow}, DifferNet~\cite{rudolph2021same}    \\ 
                         \cline{3-4} 
                                              &                                    & Memory bank                & \tab{PaDiM}~\cite{defard2021padim}, \tab{PatchCore}~\cite{roth2022towards},  \tab{SPADE}~\cite{cohen2020sub}, \tab{CFA}~\cite{lee2022cfa}, SOMAD~\cite{li2021anomaly}, \cite{kim2021semi} \\ \cline{3-4} 
                                                &                                    & Teacher-student             & \tab{RD4AD}~\cite{Deng2022AnomalyDV}, \tab{STPM}~\cite{Wang2021StudentTeacherFP}, \cite{yamada2021reconstruction}, \cite{bergmann2020uninformed}, \cite{salehi2021multiresolution}            \\ 
                         \cline{3-4} 
                                    &                                    & One-class classification   & \tab{CutPaste}~\cite{li2021cutpaste}, PANDA~\cite{Reiss2021PANDAAP}, DROC\cite{sohn2020learning}, MOCCA~\cite{massoli2021mocca},PatchSVDD~\cite{yi2020patch}, SE-SVDD~\cite{hu2021semantic}, \cite{sauter2021defect}          \\
                         
                         \cline{2-4} 
                                       & \multirow{2}{*}{Reconstruction}    & External data usage         & \tab{DREAM}~\cite{zavrtanik2021draem}, DSR~\cite{zavrtanik2022dsr}, MSTUnet\cite{jiang2022masked}, DFR~\cite{yang2020dfr}            \\ \cline{3-4} 
                          &                                    & Internal data usage only & \tab{FAVAE}~\cite{Dehaene2020AnomalyLB}, NSA~\cite{schluter2021self}, RIAD~\cite{zavrtanik2021reconstruction}, SCADN~\cite{yan2021learning}, InTra~\cite{pirnay2022inpainting}, \cite{bergmann2018improving}, \cite{hou2021divide}, \cite{liu2020towards}, \cite{collin2021improved}, \cite{yan2021unsupervised}, \cite{dehaene2019iterative}             \\ \cline{1-4} 
                          
\multicolumn{3}{l|}{\textbf{Fully supervised}}               &  \tab{DRA}~\cite{ding2022catching}, \tab{DevNet}~\cite{pang2021explainable}, \tab{PRN}~\cite{zhang2023prototypical}, \tab{BGAD}~\cite{yao2023explicit}, \tab{SemiREST}~\cite{li2023efficient}, FCDD~\cite{liznerski2020explainable}, SPD~\cite{zou2022spot}, CAVGA~\cite{venkataramanan2020attention}, \cite{chu2020neural}                   \\ \cline{1-4} 
             
\multicolumn{3}{l|}{\textbf{Few-shot}}                 &        \tab{RegAD}~\cite{huang2022registration}, RFS~\cite{kamoona2021anomaly}, \cite{sheynin2021hierarchical}            \\ \cline{1-4} 
                   
\multicolumn{3}{l|}{\textbf{Noisy label}}               &   \tab{IGD}~\cite{Chen2022DeepOC}, LOE~\cite{qiu2022latent},  TrustMAE~\cite{tan2021trustmae}, SROC \cite{cordier2022data}, SRR \cite{yoon2021self}, CPCAD~\cite{de2021contrastive}    \\ \cline{1-4} 

\multicolumn{3}{l|}{\textbf{Continual}}            &             \tab{DNE}~\cite{li2022towards}      \\ \hline
\end{tabular}
}
\label{tab:IAD_alg}
\end{table*}

\subsection{Implementation Datails}
\label{sec:baseline_models}

\subsubsection{Baseline Methods}
Table~\ref{tab:IAD_alg} lists 16 IAD algorithms (marked in \tab{purple}). The criteria for selecting algorithms to be implemented for IM-IAD are that the algorithms should be representative in terms of supervision level (fully supervised and unsupervised), noise-resilient capabilities, the facility of data-efficient adaptation (few shot), and the capacity to overcome catastrophic forgetting. Since most of them achieve state-of-the-art (SOTA) performance on the majority of industrial image datasets, they are referred to as vanilla methods and compared in the IM setting.

\subsubsection{Datasets}
To perform comprehensive ablation studies, we employ seven public datasets in the IM-IAD, including MVTec AD~\cite{Bergmann2019MVTecA,bergmann2021mvtecIJCV}, MVTec LOCO-AD~\cite{bergmann2022beyond} MPDD~\cite{jezek2021deep}, BTAD~\cite{mishra2021vt}, VisA~\cite{zou2022spot}, MTD~\cite{huang2020surface}, and DAGM~\cite{DAGMGNSS2077}. Table~\ref{tab:dataset_diff} provides an overview of these datasets, including the number of samples (normal and abnormal samples), the number of classes, the types of anomalies, and the resolution of the image. Pixel-level annotations are available for all datasets. Note that DAGM is a synthetic dataset, MVTec LOCO-AD proposes logical IAD, and VisA proposes multi-instance IAD.

\begin{table}[ht]
\centering
\caption{Comparison of datasets in IM-IAD.}
\renewcommand{\arraystretch}{1.3}
\resizebox{1\linewidth}{!}{
\huge
\begin{tabular}{l|rr|rr|rr}
    \hline
     \rowcolor{blue!5}   & \multicolumn{2}{c|}{\textbf{Sample Number}} & \multicolumn{2}{c|}{\textbf{Classes}} & \multicolumn{2}{c}{\textbf{Image Resolution}} \\\cline{2-7}
     \rowcolor{blue!5}   \textbf{Dataset}            &  \textbf{Normal} & \textbf{Anomaly} & \textbf{Anomaly Type} & \textbf{Object} & \textbf{Min} & \textbf{Max} \\\hline
    MVTec AD~\cite{Bergmann2019MVTecA,bergmann2021mvtecIJCV}      & 4,096           & 1,258            & 73              & 15             & 700             & 1,024            \\
    MVTec LOCO-AD~\cite{bergmann2022beyond} & 2,651           & 993             & 89              & 5              & 850             & 1,700            \\
    MPDD~\cite{jezek2021deep}          & 1,064           & 282             & 5               & 1              & 1,024            & 1,024            \\
    BTAD~\cite{mishra2021vt}          &       2,250         &      580           & 3               & 3              &   600              &     1,600            \\
    MTD~\cite{huang2020surface}           & 952               &    392             &       5          &        1        &    113             &  491               \\
    VisA~\cite{zou2022spot}         & 10,621          & 1,200            & 78              & 12             &   960              &   1,562              \\ 
    DAGM~\cite{DAGMGNSS2077}          &    15,000            &       2,100          &         10        &       10         &            512     &    512             \\
\hline
\end{tabular}
}
\label{tab:dataset_diff}
\end{table}

\subsubsection{Evaluation Metrics}
In terms of structural anomalies, we employ Area Under the Receiver Operating Characteristics (AU-ROC/AUC), Area Under Precision-Recall (AUPR/AP), and PRO~\cite{bergmann2021mvtecIJCV} to evaluate the abilities of anomaly localization. Regarding logical anomalies, we adopt sPRO~\cite{bergmann2021mvtec} to measure the ability of logical defect detection. Additionally, we use the Forgetting Measure (FM)~\cite{Chaudhry2018RiemannianWF} to assess the ability to resist catastrophic forgetting. The relevant formulas are shown in Table~\ref{tab:metric-summary}.

\subsubsection{Hyperparameters}
Table~\ref{tab:experiment_detail} shows the hyperparameter settings of IM-IAD, including training epochs, batch size, image size, and learning rate, respectively. We share the source codes on the website: \href{https://github.com/M-3LAB/open-iad}{https://github.com/M-3LAB/open-iad}. 

\begin{table*}[ht]
\centering
\caption{A summary of metrics used in IM-IAD.}
\renewcommand{\arraystretch}{1.3}
\resizebox{\textwidth}{!}{
\begin{tabular}{p{5cm}p{1cm}p{5cm}p{8cm}}
    \hline
     \rowcolor{blue!5} \textbf{Metric}  & \textbf{Better}     & \textbf{Formula}  & \textbf{Remarks / Usage}     \\ \hline
    Precision (P) & $\uparrow$ & $P = TP / (TP+FP)$ & True Positive (TP), False Positive (FP) \\
    Recall (R)  & $\uparrow$ & $R =TP / (TP+FN) $ & False Negative (FN)\\
    True Positive Rate (TPR)  &   $\uparrow$ & $TPR=TP/(TP+FN)$ & Classification \\
    False Positive Rate (FPR)  &  $\downarrow$ & $FPR=FP/FP+TN)$  & True Negative (TN) \\
    \makecell[l]{ Area Under the Receiver Operating \\ Characteristic curve (AU-ROC)~\cite{bergmann2021mvtecIJCV}}  &  $\uparrow$ & $\int_{0}^{1}(TPR)\,{\rm d}(FPR) $   & Classification              \\
    Area Under Precision-Recall (AU-PR)~\cite{bergmann2021mvtecIJCV}  &  $\uparrow$    & $ \int_{0}^{1}(P)\,{\rm d}(R)$     & Localization, Segmentation \\
    Per-Region Overlap (PRO)~\cite{bergmann2021mvtecIJCV}  &   $\uparrow$      &  $PRO = \frac{1}{N} \sum \limits_i \sum \limits_k \frac{P_i \cap C_{i, k}}{C_{i, k}} $      & \makecell[l]{Total ground-truth number (N), Predicted abnormal pixels (P),\\ Defect ground-truth regions (C), Segmentation}      \\
    Saturated Per-Region Overlap (sPRO)~\cite{bergmann2021mvtec} &   $\uparrow$         & $sPRO(P)=\frac{1}{m} \sum \limits_{i=1}^{m} \min(\frac{A_i \cap P}{s_i}, 1)$      & \makecell[l]{Total ground-truth number (m), Predicted abnormal pixels (P),\\ Defect ground-truth regions (A), Corresponding saturation\\ thresholds (s), Segmentation}    \\
    Forgetting Measure (FM)~\cite{Chaudhry2018RiemannianWF}  & $\downarrow$ & $FM_j^k=\max\limits_{l\in\{1,...,k-1\}}\mathbf{T}_{l,j}-\mathbf{T}_{k,j}$ & 
    Task (T), Number of tasks (k), Task to be evaluated (j) \\ \hline
\end{tabular}
}
\label{tab:metric-summary}
\end{table*}

\begin{table*}[th]
\caption{Details of experimental settings.}
\renewcommand{\arraystretch}{1.3}
\resizebox{\linewidth}{!}{
\begin{tabular}{l|cccccccccccccc}
\hline
   \rowcolor{blue!5}   \textbf{Method}         & \textbf{CFA}   & \textbf{CSFlow} & \textbf{CutPaste} & \textbf{DNE}    & \textbf{DRAEM}  & \textbf{FastFlow} & \textbf{FAVAE}   & \textbf{IGD}    & \textbf{PaDiM}            & \textbf{PatchCore}        & \textbf{RegAD}  & \textbf{RD4AD} & \textbf{SPADE}            & \textbf{STPM} \\ \hline
Training Epochs & 50    & 240    & 256      & 50     & 700    & 500      & 100     & 256    & 1                & 1                & 50     & 200   & 1                & 100  \\
Batch Size     & 4     & 16     & 32       & 32     & 8      & 32       & 64      & 16     & 32               & 2                & 32     & 8     & 8                & 8    \\
Image Size     & 256   & 768    & 224      & 224    & 256    & 256      & 256     & 256    & 256              & 256              & 224    & 256   & 256              & 256  \\
Learning Rate  & 0.001 & 0.0002 & 0.0001   & 0.0001 & 0.0001 & 0.001    & 0.00001 & 0.0001 & \textbackslash{} & \textbackslash{} & 0.0001 & 0.005 & \textbackslash{} & 0.4 \\ \hline
\end{tabular}}
\label{tab:experiment_detail}
\end{table*}

\subsection{Uniform Perspective} 

The proposed IM-IAD bridges the connection of different existing settings, as there is no uniform evaluation benchmark. 

1) From the perspective of training datasets, the type and number of training samples are the main differences between the five settings shown in Fig.~\ref{fig:fewshot_ad_paradigm}. It should be noted that unsupervised IAD algorithms only use normal datasets for training. On the contrary, fully supervised IAD and noisy IAD incorporate a limited quantity of abnormal samples during the training phase. Few-shot IAD employs a limited number of normal samples ($\leq$ 8) for training.

2) From the perspective of applications, IM-IAD are designed to accommodate different scenarios in the real-production line. Since collecting many abnormal samples for training is difficult, unsupervised IAD are designed for a general case via just using normal training samples. Few-shot IAD aims to solve the challenge of cold start in a single assembly line scenario, where normal samples are limited. The goal of continual IAD is to address catastrophic forgetting that occurs when IAD models are integrated into the recycling assembly line. Noisy IAD tries to eliminate the side effects resulting from contaminated training data. Full supervised IAD aims to improve the efficiency of abnormal sample utilization because labelling anomalies is expensive.

\subsection{Open Challenges}
Regarding the setting in IM-IAD, the following is a summary of the challenging issues that need to be investigated.

1) The remaining challenges for fully supervised IAD~\cite{ding2022catching, li2023efficient} described in Fig.~\ref{fig:fewshot_ad_paradigm}(c) is how to effectively use the guidance of limited abnormal data and a large number of normal samples to detect the anomalies.

2) For the few-shot IAD shown in Fig.~\ref{fig:fewshot_ad_paradigm}(d), we aim to detect anomalies in the test set using a small number of normal or abnormal images in the training set~\cite{xie2023pushing}. The main obstacles are: i) In a few-shot setting, the training dataset for each category contains only normal samples, meaning there are no annotations at the image or pixel level. ii) Few normal samples of the training set are accessible. In the setting we propose, there are less than eight training samples. 

3) We attempt to detect abnormal samples and identify anomalies given a target category for IAD in the presence of noise~\cite{jiang2022softpatch} described in Fig.~\ref{fig:fewshot_ad_paradigm}(e). For example, clean training data is assumed to consist exclusively of normal samples. While contaminated training data contains noisy samples incorrectly labelled as normal, \textit{i.e.}, label flipping. Anormal samples are easily mislabeled as normal because the anomalies are too small to identify. The most significant barriers are summarized as follows. i) Each category's training set contains noisy data that could easily confuse the decision threshold of IAD algorithms. ii) There is a large amount of noisy data. Here, the percentage of anomalous samples in the training set ranges from 5\% to 20\%. Hence, noisy IAD aims to assess the resilience of current unsupervised IAD methods in the presence of contaminated data. 

4) For the continual IAD presented in Fig.~\ref{fig:fewshot_ad_paradigm}(f), the greatest challenge is that IAD algorithms may suffer from catastrophic forgetting when they have completed training on the new category dataset. In real-world applications, a single assembly line typically accommodates a substantial amount of workpiece, often containing thousands. Deploying numerous IAD models on a single assembly line is unfeasible due to the high expenses associated with maintenance. Furthermore, a large part of the assembly line process involves recycling. Hence, industrial deployment requires IAD models to overcome catastrophic forgetting.

\begin{figure*}[th]
\centering 
\subfigure[MVTec AD]{
	\label{fig:efficiency_mvtec2d}
    \includegraphics[width=0.45\textwidth]{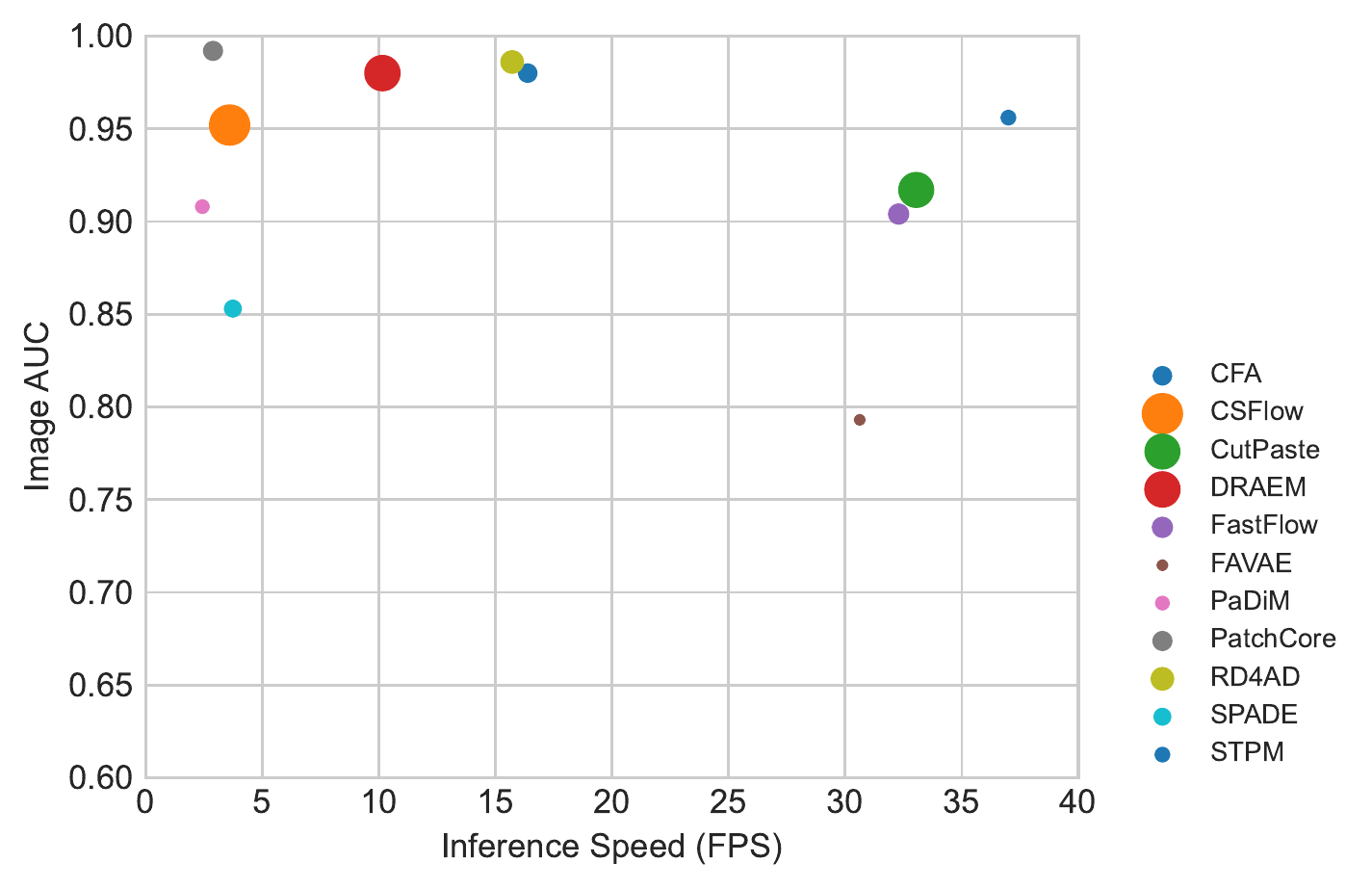}
    }
\subfigure[MVTec LOCO-AD]{
    \label{fig:efficiency_mvtecloco}
    \includegraphics[width=0.45\textwidth]{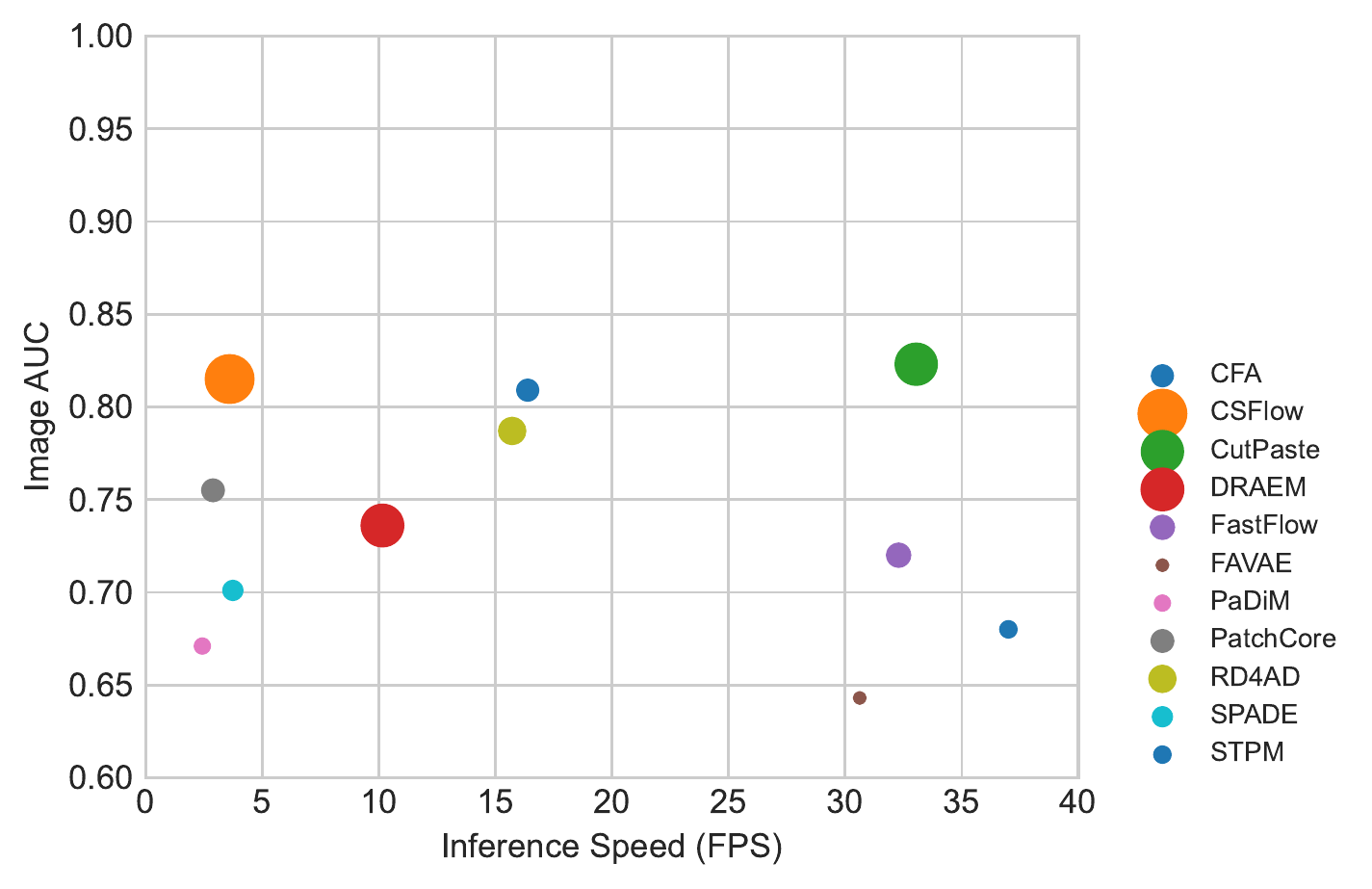}
    }
\caption{Visualization of vanilla IAD algorithms on Image AUC $\uparrow$, inference time and GPU memory under MVTec AD and LOCO-AD. The Y-axis denotes the performance of the IAD model. The X-axis refers to the inference time taken for each image. The size of the circle denotes the GPU memory consumption of the IAD model during the test phase, where the small one is better.}
\label{fig:efficiency}
\end{figure*}

\section{Results and Discussions}~\label{sec:benchmark_result}

This section explores existing algorithms and discusses the critical aspects of the proposed uniform settings. Each part describes experimental facilities, analyzes results, and provides other challenges and future directions.

\subsection{Overall Comparisons}

\textbf{Settings.} The vanilla methods are presented in Table~\ref{tab:IAD_alg}. The unsupervised setting is described in Sec.~\ref{sec:definition}-1.

\begin{table*}[th]
\caption{Comparison of vanilla IAD algorithms for 7 datasets using 5 metrics. We report sPRO with a 0.05 integration parameter. The best and second-best results are marked in red and blue, respectively.}
\renewcommand{\arraystretch}{1.2}
\resizebox{\linewidth}{!}{
\begin{tabular}{l|l|lllllllllll}
\hline
 \rowcolor{blue!5} \textbf{Dataset}                         & \textbf{Metric} $\uparrow$               & \textbf{CFA}                          & \textbf{CS-Flow }                     & \textbf{CutPaste}                     & \textbf{DRAEM}                        & \textbf{FastFlow} & \textbf{FAVAE}                        & \textbf{PaDiM}                        & \textbf{PatchCore}                    & \textbf{RD4AD}                        & \textbf{SPADE}                        & \textbf{STPM}                         \\ \hline
                                & Image AUC            & 0.981                        & 0.952                        & 0.918                        & 0.981                        & 0.905    & 0.793                        & 0.908                        & {\color[HTML]{FE0000} 0.992} & {\color[HTML]{0000FF} 0.986} & 0.854                        & 0.924                        \\
                                & Image AP             & 0.993                        & 0.975                        & 0.965                        & 0.990                        & 0.945    & 0.913                        & 0.954                        & {\color[HTML]{FE0000} 0.998} & {\color[HTML]{0000FF} 0.995} & 0.940                        & 0.957                        \\
                                & Pixel AUC            & 0.971                        & --                            & --                            & {\color[HTML]{0000FF} 0.975} & 0.955    & 0.889                        & 0.966                        & 0.994                        & {\color[HTML]{FE0000} 0.978} & 0.955                        & 0.954                        \\
                                & Pixel AP             & 0.538                        & --                            &  --                           & {\color[HTML]{FE0000} 0.689} & 0.398    & 0.307                        & 0.452                        & 0.561                        & {\color[HTML]{0000FF} 0.580} & 0.471                        & 0.518                        \\
\multirow{-5}{*}{MVTec AD}      & Pixel PRO            & 0.898                        &        --                     &     --                        & 0.921                        & 0.856    & 0.749                        & 0.913                        & {\color[HTML]{FE0000} 0.943} & {\color[HTML]{0000FF} 0.939} & 0.895                        & 0.879                        \\ \hline
                                & Image AUC            & 0.814                        & 0.814                        & 0.734                        & 0.798                        & 0.639    & 0.623                        & 0.780                        & {\color[HTML]{0000FF} 0.835} & {\color[HTML]{FE0000} 0.867} & 0.687                        & 0.679                        \\
                                & Image AP             & 0.944                        & 0.942                        & 0.915                        & 0.933                        & 0.866    & 0.873                        & 0.927                        & {\color[HTML]{0000FF} 0.948} & {\color[HTML]{FE0000} 0.958} & 0.890                        & 0.891                        \\
                                & Pixel AUC            & 0.908                        &      --                       &     --                        & 0.942                        & 0.796    & 0.944                        & {\color[HTML]{0000FF} 0.987} & {\color[HTML]{FE0000} 0.990} & 0.971                        & 0.971                        & 0.848                        \\
                                & Pixel AP             & 0.219                        &    --                         &    --                         & 0.209                        & 0.053    & 0.099                        & 0.149                        & 0.150                        & {\color[HTML]{FE0000} 0.342} & {\color[HTML]{0000FF} 0.261} & 0.164                        \\
\multirow{-6}{*}{MVTec LOCO-AD} & Mean sPRO       & {\color[HTML]{0000FF} 0.581} &   --                          &  --                           & 0.426                        & 0.357    & 0.446                        & 0.521                        & 0.343                        & {\color[HTML]{FE0000} 0.637} & 0.520                        & 0.428                        \\ \hline
                                & Image AUC            & 0.923                        & {\color[HTML]{FE0000} 0.973} & 0.771                        & 0.941                        & 0.887    & 0.570                        & 0.706                        & {\color[HTML]{0000FF} 0.948} & 0.927                        & 0.784                        & 0.876                        \\
                                & Image AP             & 0.922                        & {\color[HTML]{0000FF} 0.968} & 0.800                        & 0.961                        & 0.881    & 0.705                        & 0.784                        & {\color[HTML]{FE0000} 0.970} & 0.953                        & 0.815                        & 0.914                        \\
                                & Pixel AUC            & 0.948                        &         --                    &     --                        & 0.918                        & 0.808    & 0.906                        & 0.955                        & {\color[HTML]{FE0000} 0.990} & {\color[HTML]{0000FF} 0.987} & 0.982                        & 0.981                        \\
                                & Pixel AP             & 0.283                        &   --                          &   --                          & 0.288                        & 0.115    & 0.088                        & 0.155                        & {\color[HTML]{0000FF} 0.432} & {\color[HTML]{FE0000} 0.455} & 0.342                        & 0.354                        \\
\multirow{-5}{*}{MPDD}          & Pixel PRO            & 0.832                        &    --                         &     --                        & 0.781                        & 0.498    & 0.706                        & 0.848                        & {\color[HTML]{0000FF} 0.939} & {\color[HTML]{FE0000} 0.953} & 0.926                        & {\color[HTML]{0000FF} 0.939} \\ \hline
                                & Image AUC            & 0.938                        & 0.936                        & 0.917                        & 0.895                        & 0.919    & 0.923                        & {\color[HTML]{FE0000} 0.965} & {\color[HTML]{0000FF} 0.947} & 0.937                        & 0.904                        & 0.918                        \\
                                & Image AP             & 0.980                        & 0.890                        & 0.953                        & 0.974                        & 0.867    & {\color[HTML]{0000FF} 0.986} & 0.976                        & {\color[HTML]{FE0000} 0.989} & 0.985                        & 0.974                        & 0.962                        \\
                                & Pixel AUC            & 0.959                        &    --                         &    --                         & 0.874                        & 0.965    & 0.949                        & {\color[HTML]{0000FF} 0.977} & {\color[HTML]{FE0000} 0.978} & 0.958                        & 0.950                        & 0.937                        \\
                                & Pixel AP             & 0.517                        &       --                      &    --                         & 0.159                        & 0.379    & 0.349                        & {\color[HTML]{FE0000} 0.535} & {\color[HTML]{0000FF} 0.520} & 0.517                        & 0.441                        & 0.401                        \\
\multirow{-5}{*}{BTAD}          & Pixel PRO            & 0.702                        &   --                          &  --                           & 0.629                        & 0.725    & 0.713                        & {\color[HTML]{FE0000} 0.798} & {\color[HTML]{0000FF} 0.752} & 0.723                        & 0.745                        & 0.667                        \\ \hline
                                & Image AUC            & {\color[HTML]{0000FF} 0.913} & 0.887                        & 0.830                        & 0.782                        & 0.891    & 0.795                        & 0.885                        & {\color[HTML]{FE0000} 0.975} & 0.884                        & 0.868                        & 0.729                        \\
                                & Image AP             & {\color[HTML]{0000FF} 0.959} & 0.945                        & 0.912                        & 0.885                        & 0.947    & 0.867                        & 0.944                        & {\color[HTML]{FE0000} 0.988} & 0.947                        & 0.928                        & 0.847                        \\
                                & Pixel AUC            & 0.731                        & --                            &   --                          & 0.660                        & 0.710    & 0.735                        & {\color[HTML]{0000FF}0.768}                        & {\color[HTML]{FE0000} 0.836} & 0.693                        &  0.742 & 0.642                        \\
                                & Pixel AP             & 0.246                        &     --                        &      --                       & 0.148                        & 0.172    & 0.120                        & {\color[HTML]{FE0000} 0.768} & {\color[HTML]{0000FF} 0.303} & 0.218                        & 0.123                        & 0.102                        \\
\multirow{-5}{*}{MTD}           & Pixel PRO            & 0.528                        &   --                          &  --                           & 0.541                        & 0.568    & 0.632                        & {\color[HTML]{FE0000} 0.798} & {\color[HTML]{0000FF} 0.686} & 0.623                        & 0.627                        & 0.478                        \\ \hline
                                & Image AUC            & 0.920                        & 0.744                        & 0.819                        & 0.887                        & 0.822    & 0.803                        & 0.891                        & {\color[HTML]{0000FF} 0.951} & {\color[HTML]{FE0000} 0.960} & 0.821                        & 0.833                        \\
                                & Image AP             & 0.935                        & 0.787                        & 0.848                        & 0.905                        & 0.843    & 0.843                        & 0.895                        & {\color[HTML]{0000FF} 0.962} & {\color[HTML]{FE0000} 0.965} & 0.847                        & 0.873                        \\
                                & Pixel AUC            & 0.843                        &           --                  &   --                          & 0.935                        & 0.882    & 0.880                        & {\color[HTML]{0000FF} 0.981} & {\color[HTML]{FE0000} 0.988} & 0.901                        & 0.856                        & 0.834                        \\
                                & Pixel AP             & 0.268                        &       --                      &   --                          & 0.265                        & 0.156    & 0.213                        & {\color[HTML]{0000FF} 0.309} & {\color[HTML]{FE0000} 0.401} & 0.277                        & 0.215                        & 0.169                        \\
\multirow{-5}{*}{VisA}          & Pixel PRO            & 0.551                        &     --                        &     --                        & 0.724                        & 0.598    & 0.679                        & {\color[HTML]{0000FF} 0.859} & {\color[HTML]{FE0000} 0.912} & 0.709                        & 0.659                        & 0.620                        \\ \hline
                                & Image AUC            & {\color[HTML]{0000FF} 0.948} & 0.752                        & 0.839                        & 0.908                        & 0.874    & 0.695                        & 0.940                        & 0.936                        & {\color[HTML]{FE0000} 0.958} & 0.714                        & 0.739                        \\
                                & Image AP             & {\color[HTML]{0000FF} 0.878} & 0.781                        & 0.680                        & 0.790                        & 0.699    & 0.376                        & 0.811                        & 0.826                        & {\color[HTML]{FE0000} 0.901} & 0.392                        & 0.498                        \\
                                & Pixel AUC            & 0.942                        &     --                        &  --                           & 0.868                        & 0.911    & 0.804                        & 0.961 & {\color[HTML]{0000FF}0.967}                        & {\color[HTML]{FE0000} 0.975} & 0.880                        & 0.859                        \\
                                & Pixel AP             & 0.495                        &  --                           &    --                         & 0.306                        & 0.342    & 0.170                        & 0.492                        & {\color[HTML]{0000FF} 0.517} & {\color[HTML]{FE0000} 0.534} & 0.133                        & 0.151                        \\
\multirow{-5}{*}{DAGM}          & Pixel PRO            & 0.870                        & --                            & --                            & 0.710                        & 0.799    & 0.600                        & {\color[HTML]{0000FF}0.906}                        &  0.893 & {\color[HTML]{FE0000} 0.930} & 0.707                        & 0.668                        \\ \cline{2-13} \hline
\end{tabular}}
\label{tab:vanilla_performance}
\end{table*}

\textbf{Discussions.} 
The statistical results of Table~\ref{tab:vanilla_performance} indicate that there is no universal winner for all datasets. Furthermore, Fig.~\ref{fig:efficiency_mvtec2d} and Fig.~\ref{fig:efficiency_mvtecloco} show that there are no dominant solutions to GPU accuracy, inference speed, and memory. Specifically, Table~\ref{tab:vanilla_performance} indicates that PatchCore, one of the most advanced memory bank based methods, performs better on MVTec AD than on MVTec LOCO-AD. Because PatchCore architectures specialize in structural anomalies, not logical anomalies, the main differences between MVTec AD and MVTec LOCO-AD are the types of anomalies.

\textbf{Logical Anomaly Definition.}
MVTec LOCO-AD consists of structural and logical anomalies. Logical anomalies are not dents or scratches, but are caused by displacement or missing parts. However, MVTec AD only has a structural abnormality. For visualization results, Fig.~\ref{fig:data_selection} shows the restrictions of each IAD model to different types of anomalies. For example, the fifth-column images are screw bags, and their defects are logical anomalies. In particular, each box of non-defective screw bags contains exactly \textit{one pushpin}. However, the defective screw bag contains two pushpins in the upper right corner of the box. The patchcore heatmap cannot accurately represent the anomaly on the right, \textit{i.e.}, whereas RD4AD~\cite{Deng2022AnomalyDV} can precisely identify the logical anomaly on the top right.

\textbf{Memory Usage and Inference Speed.} 
 From Fig.~\ref{fig:efficiency}, it is clear that there are no dominant IAD methods in terms of accuracy, memory usage and inference time. PatchCore achieves SOTA performance in image AUC, but does not take advantage of memory usage and inference time. In practical scenarios, the use of memory and the speed of inference must be fully taken into account. Therefore, current advanced IAD algorithms cannot meet the requirements of IM.

\revised{
\textbf{Image-Level or Pixel-Level Evaluation?} 
Researchers commonly use image-level and pixel-level metrics to evaluate the classification performance of IAD algorithms. In practice, the image-level metric is used to judge whether the whole product is abnormal, while the pixel-level metric indicates anomaly localization performance. To be specific, the pixel-level metric can evaluate the degree of defection, which is strongly associated with the price of products. A lower degree of defects implies a higher price, or vice versa. According to Table~\ref{tab:vanilla_performance}, specific IAD methods, like PatchCore, perform well on image AUROC but poorly on pixel AP, or vice versa. These two types of metrics signify distinct capabilities of IAD algorithms, and both are very important for IM. \textbf{Therefore, it is imperative to develop innovative IAD algorithms that exhibit exceptional performance in terms of both image-level metrics and pixel-level metrics. }
}

\textbf{Challenges.} 
How do we use the types of visual anomalies to select and design unsupervised algorithms effectively? Algorithm selection based on anomalous types is crucial, but there is a lack of research in this area. Since product line specialists can provide information on the type of anomalies, it is appropriate to be aware of the kind of visual anomalies in advance. In other words, knowledge of anomalous types in an unsupervised algorithm can be considered supervision information. Therefore, algorithm designers should consider abnormal types when developing algorithms.

\subsection{Role of Global Features in Logical IAD}

\textbf{Settings.} We benchmark the vanilla unsupervised IAD algorithms on the MVTec LOCO-AD dataset. In addition, we have re-implemented the baseline logical AD method, GCAD~\cite{bergmann2022beyond}.

\textbf{Discussions.} 
According to Table~\ref{tab:loco_comparsion}, the baseline method GCAD~\cite{bergmann2022beyond} is superior to all unsupervised IAD methods. The key idea of GCAD~\cite{bergmann2022beyond} is to encode each pixel's feature descriptor through a bottleneck architecture into a global feature. Existing unsupervised IAD approaches have the disadvantage that their architectures are not optimized to acquire global features.

\begin{figure}[th]
    \centering
    \includegraphics[width=1\linewidth]{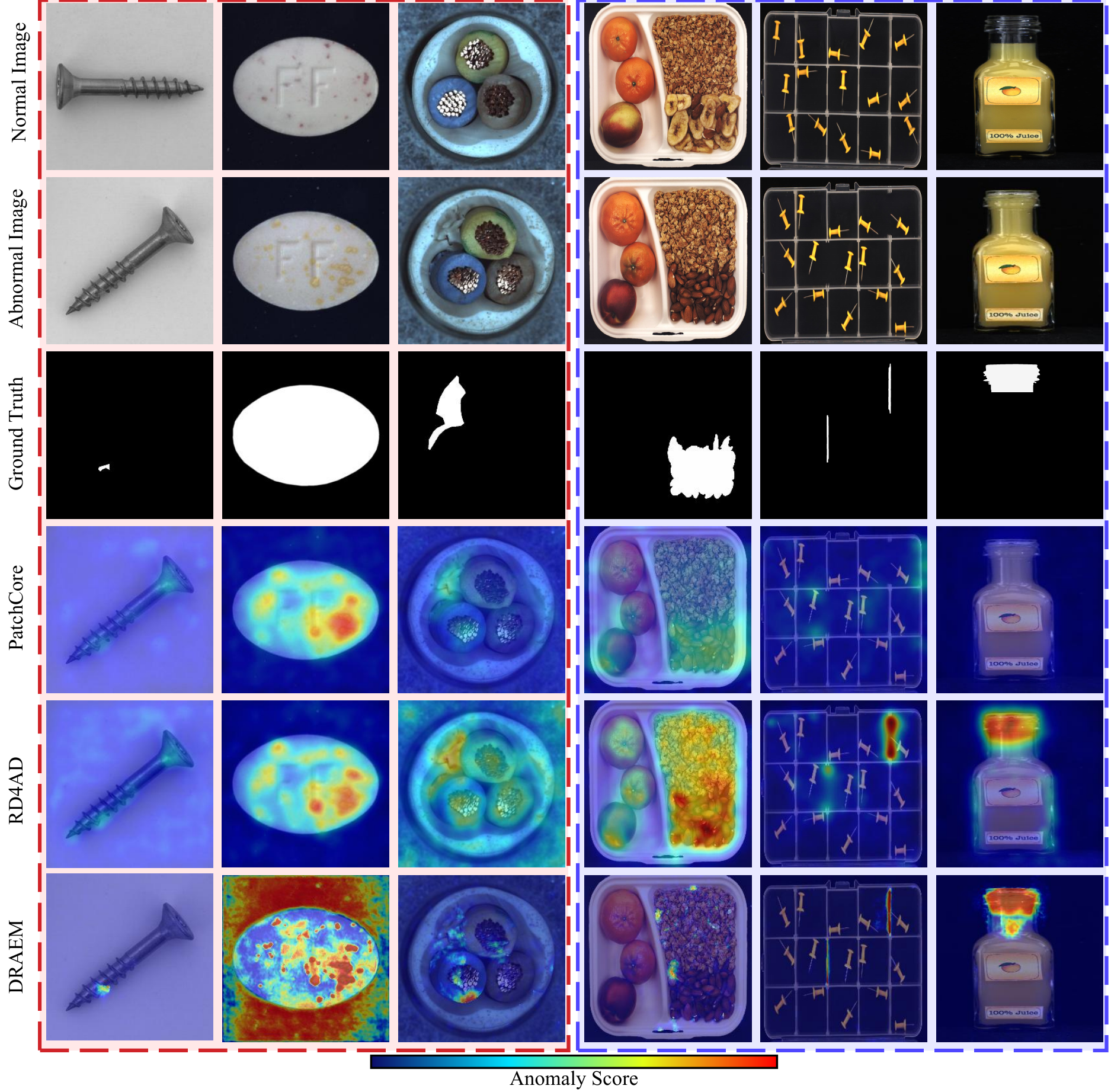}
    \caption{Visualization of the representative vanilla IAD algorithms. The three columns on the left (marked in \textcolor{red}{red}) show structural anomalies, while the three columns on the right (marked in \textcolor{blue}{blue}) show logical anomalies. The first row indicates the training images, where all training images are normal. The second row denotes a test abnormal image and the third row reals the anomalies of the above abnormal image. Lastly, the fourth to sixth row presents the heat map of PatchCore, RD4AD and DRAEM, respectively. }
\label{fig:data_selection}		
\end{figure}

\begin{table}[th]
\centering
\caption{Benchmark on MVTec LOCO-AD in terms of logical anomalies, structural anomalies and their mean value. The best and second-best results are marked in red and blue, respectively.}
\renewcommand{\arraystretch}{1.2}
\resizebox{1\linewidth}{!}{
\begin{tabular}{l|lll|lll}
\hline
\rowcolor{blue!5}  & \multicolumn{3}{c|}{\textbf{Image AUROC $\uparrow$}} & \multicolumn{3}{c}{\textbf{Pixel sPRO $\uparrow$}} \\\cline{2-7}
\rowcolor{blue!5} \textbf{Method}  & \textbf{Logical}   & \textbf{Structural}  & \textbf{Mean}   & \textbf{Logical}  & \textbf{Strctural}  & \textbf{Mean}   \\ \hline
PatchCore         & 0.690     & 0.820       & 0.755  & 0.340    & 0.345      & 0.343  \\
CFA               & 0.768     & 0.851       & 0.809  & {\color[HTML]{0000FF}0.536}    & 0.625      & 0.581  \\
SPADE             & 0.653     & 0.749       & 0.701  & 0.430    & 0.609      & 0.520  \\
PaDiM             & 0.637     & 0.705       & 0.671  & 0.517    & 0.525      & 0.521  \\
RD4AD             & 0.694     & {\color[HTML]{FE0000} 0.880}       & 0.787  & 0.497    & {\color[HTML]{FE0000} 0.777}      & {\color[HTML]{0000FF} 0.637}  \\
STPM              & 0.597     & 0.763       & 0.680  & 0.328    & 0.529      & 0.428  \\
CutPaste          & 0.779     & {\color[HTML]{0000FF} 0.867}       & {\color[HTML]{0000FF} 0.823}  & --       & --       & --      \\
CSFlow            & {\color[HTML]{0000FF}0.783}     & 0.847       & 0.815  & --        & --          & --      \\
FastFlow          & 0.727     & 0.712       & 0.720  & 0.359    & 0.356      & 0.357  \\
DRAEM             & 0.728     & 0.744       & 0.736  & 0.454    & 0.398      & 0.426  \\
FAVAE             & 0.659     & 0.628       & 0.643  & 0.501    & 0.392      & 0.446  \\
GCAD              & {\color[HTML]{FE0000}0.860}     & 0.806       & {\color[HTML]{FE0000}0.833}  & {\color[HTML]{FE0000}0.711}    & {\color[HTML]{0000FF}0.692}      &{\color[HTML]{FE0000} 0.701}      \\\hline
\end{tabular}}

\label{tab:loco_comparsion}
\end{table}

\textbf{Challenges.} 
Global feature extraction is crucial to achieving a high detection performance for logical IAD tasks. The statistical results presented in Table~\ref{tab:loco_comparsion} highlight the significance of global anomaly feature extraction. Recent network architectures such as Transformer~\cite{Dosovitskiy2020AnII} and Normalizing Flow~\cite{Kobyzev2019NormalizingFI} focus on long-distance feature extraction, which makes it easier to detect logical anomalies. The results in Table~\ref{tab:loco_comparsion} show that CSFlow~\cite{rudolph2022fully}, based on Normalizing Flow, achieves the second-best performance regarding logical anomalies, indicating its potential. Additionally, the bottleneck architecture is another feasible approach to capturing global features. As demonstrated by the heat map on the logical anomaly dataset in Fig.~\ref{fig:data_selection}, the bottleneck design of RD4AD~\cite{Deng2022AnomalyDV} is capable of extracting global features.

\subsection{Abnormal Data for Fully Supervised IAD}

\textbf{Settings.} We first benchmark fully supervised IAD methods according to the setting in~\cite{pang2021explainable,chen2022deep,li2023efficient}. According to the definition of fully supervised IAD in Sec.~\ref{sec:definition}-2, we have set the number of abnormal training samples to 10. Here, we make comparisons with PRN~\cite{zhang2023prototypical}, BGAD~\cite{yao2023explicit}, DevNet~\cite{pang2021explainable}, DRA~\cite{ding2022catching}, SemiREST~\cite{li2023efficient}, and PatchCore~\cite{roth2022towards}.

\begin{table*}[th]
\centering
\caption{Performance on the fully-supervised setting. The best and second-best results are marked in red and blue, respectively.}
\renewcommand{\arraystretch}{1.2}
\resizebox{\linewidth}{!}{
\begin{tabular}{l|lllllllllllllllll}
\hline
  \rowcolor{blue!5}      \textbf{Method}  & \textbf{Metric} $\uparrow$  & \textbf{Bottle} & \textbf{Cable} & \textbf{Capsule} & \textbf{Carpet} & \textbf{Grid} & \textbf{Hazelnut} & \textbf{Leather} & \textbf{Metal\_nut} & \textbf{Pill} & \textbf{Screw} & \textbf{Tile} & \textbf{Toothbrush} & \textbf{Transistor} & \textbf{Wood} & \textbf{Zipper} & \textbf{Mean} \\ \hline
  
& Pixel AUC                         & {\color[HTML]{0000FF} 0.994}     & {\color[HTML]{0000FF} 0.988}    & 0.985                             & 0.990                            & 0.984                          & {\color[HTML]{0000FF} 0.997}       & 0.997                             & {\color[HTML]{0000FF} 0.997}         & {\color[HTML]{FF0000} 0.995}   & 0.975                           & {\color[HTML]{0000FF} 0.996}   & {\color[HTML]{FF0000} 0.996}         & {\color[HTML]{0000FF} 0.984}         & 0.978                          & 0.988                            & 0.990                          \\
                                                                          & Pixel AP                          & {\color[HTML]{0000FF} 0.923}     & 0.789                           & {\color[HTML]{FF0000} 0.622}      & 0.820                            & 0.457                          & {\color[HTML]{FF0000} 0.938}       & 0.697                             & {\color[HTML]{0000FF} 0.980}         & {\color[HTML]{0000FF} 0.913}   & 0.449                           & {\color[HTML]{0000FF} 0.965}   & {\color[HTML]{FF0000} 0.781}         & {\color[HTML]{FF0000} 0.856}         & {\color[HTML]{0000FF} 0.826}   & 0.776                            & {\color[HTML]{0000FF} 0.786}   \\
\multirow{-3}{*}{PRN}                          & Pixel PRO                         & 0.970                            & {\color[HTML]{0000FF} 0.972}    & 0.925                             & {\color[HTML]{0000FF} 0.970}     & 0.959                          & 0.974                              & {\color[HTML]{0000FF} 0.992}      & 0.958                                & {\color[HTML]{0000FF} 0.972}   & 0.924                           & {\color[HTML]{0000FF} 0.982}   & 0.956                                & 0.948                                & 0.959                          & 0.955                            & 0.961                          \\ \hline
                                                                          & Pixel AUC                         & 0.993                            & 0.985                           & {\color[HTML]{0000FF} 0.988}      & {\color[HTML]{0000FF} 0.996}     & 0.984                          & 0.994                              & {\color[HTML]{0000FF} 0.998}      & 0.996                                & 0.995                          & {\color[HTML]{0000FF} 0.993}    & 0.993                          & {\color[HTML]{0000FF} 0.995}         & 0.979                                & {\color[HTML]{0000FF} 0.980}   & {\color[HTML]{0000FF} 0.993}     & 0.992                          \\
                                                                          & Pixel AP                          & 0.871                            & 0.814                           & 0.583                             & {\color[HTML]{0000FF} 0.832}     & {\color[HTML]{0000FF} 0.592}   & 0.824                              & {\color[HTML]{0000FF} 0.755}      & 0.973                                & {\color[HTML]{FF0000} 0.921}   & {\color[HTML]{0000FF} 0.553}    & 0.940                          & 0.713                                & 0.823                                & 0.787                          & {\color[HTML]{0000FF} 0.782}     & 0.784                          \\
\multirow{-3}{*}{BGAD}                               & Pixel PRO                         & {\color[HTML]{0000FF} 0.971}     & {\color[HTML]{FF0000} 0.977}    & {\color[HTML]{FF0000} 0.968}      & 0.989                            & {\color[HTML]{FF0000} 0.987}   & {\color[HTML]{FF0000} 0.986}       & 0.995                             & {\color[HTML]{0000FF} 0.968}         & 0.987                          & {\color[HTML]{0000FF} 0.968}    & 0.979                          & {\color[HTML]{0000FF} 0.964}         & {\color[HTML]{0000FF} 0.971}         & {\color[HTML]{0000FF} 0.968}   & {\color[HTML]{0000FF} 0.977}     & {\color[HTML]{0000FF} 0.977}   \\ \hline
                                                                          & Pixel AUC                         & 0.939                            & {\color[HTML]{0000FF} 0.888}    & 0.918                             & 0.972                            & 0.879                          & 0.911                              & 0.942                             & 0.778                                & 0.826                          & 0.603                           & 0.927                          & 0.846                                & 0.560                                & 0.864                          & 0.937                            & 0.853                          \\
                                                                          & Pixel AP                          & 0.515                            & 0.360                           & 0.155                             & 0.457                            & 0.255                          & 0.221                              & 0.081                             & 0.356                                & 0.146                          & 0.014                           & 0.523                          & 0.067                                & 0.064                                & 0.251                          & 0.196                            & 0.244                          \\
\multirow{-3}{*}{DevNet}                         & Pixel PRO                         & 0.835                            & 0.809                           & 0.836                             & 0.858                            & 0.798                          & 0.836                              & 0.885                             & 0.769                                & 0.692                          & 0.311                           & 0.789                          & 0.335                                & 0.391                                & 0.754                          & 0.813                            & 0.714                          \\ \hline
                                                                          & Pixel AUC                         & 0.913                            & 0.866                           & 0.893                             & 0.982                            & 0.860                          & 0.896                              & 0.938                             & 0.795                                & 0.845                          & 0.540                           & 0.923                          & 0.755                                & 0.791                                & 0.829                          & 0.969                            & 0.853                          \\
                                                                          & Pixel AP                          & 0.412                            & 0.347                           & 0.117                             & 0.523                            & 0.268                          & 0.225                              & 0.056                             & 0.299                                & 0.216                          & 0.050                           & 0.576                          & 0.045                                & 0.110                                & 0.227                          & 0.429                            & 0.260                          \\
\multirow{-3}{*}{DRA}                               & Pixel PRO                         & 0.776                            & 0.777                           & 0.791                             & 0.922                            & 0.715                          & 0.869                              & 0.840                             & 0.767                                & 0.770                          & 0.301                           & 0.815                          & 0.561                                & 0.490                                & 0.697                          & 0.910                            & 0.733                          \\ \hline
                                                                          & Pixel AUC                         & {\color[HTML]{FF0000} 0.995}     & {\color[HTML]{FF0000} 0.992}    & {\color[HTML]{0000FF} 0.988}      & {\color[HTML]{FF0000} 0.997}     & {\color[HTML]{FF0000} 0.994}   & {\color[HTML]{FF0000} 0.998}       & {\color[HTML]{FF0000} 0.999}      & {\color[HTML]{FF0000} 0.999}         & {\color[HTML]{0000FF} 0.993}   & {\color[HTML]{FF0000} 0.998}    & {\color[HTML]{FF0000} 0.997}   & {\color[HTML]{FF0000} 0.996}         & {\color[HTML]{FF0000} 0.986}         & {\color[HTML]{FF0000} 0.992}   & {\color[HTML]{FF0000} 0.997}     & {\color[HTML]{FF0000} 0.995}   \\
                                                                          & Pixel AP                          & {\color[HTML]{FF0000} 0.936}     & {\color[HTML]{FF0000} 0.895}    & {\color[HTML]{0000FF} 0.600}      & {\color[HTML]{FF0000} 0.891}     & {\color[HTML]{FF0000} 0.664}   & {\color[HTML]{0000FF} 0.922}       & {\color[HTML]{FF0000} 0.817}      & {\color[HTML]{FF0000} 0.991}         & 0.861                          & {\color[HTML]{FF0000} 0.721}    & {\color[HTML]{FF0000} 0.969}   & {\color[HTML]{0000FF} 0.742}         & {\color[HTML]{0000FF} 0.855}         & {\color[HTML]{FF0000} 0.887}   & {\color[HTML]{FF0000} 0.910}     & {\color[HTML]{FF0000} 0.844}   \\ 
\multirow{-3}{*}{SemiREST}                           & Pixel PRO                         & {\color[HTML]{FF0000} 0.985}     & 0.959                           & 0.970                             & {\color[HTML]{FF0000} 0.991}     & {\color[HTML]{0000FF} 0.970}   & {\color[HTML]{0000FF} 0.983}       & {\color[HTML]{FF0000} 0.997}      & {\color[HTML]{FF0000} 0.982}         & {\color[HTML]{FF0000} 0.989}   & {\color[HTML]{FF0000} 0.988}    & {\color[HTML]{FF0000} 0.989}   & {\color[HTML]{FF0000} 0.971}         & {\color[HTML]{FF0000} 0.978}         & {\color[HTML]{FF0000} 0.979}   & {\color[HTML]{FF0000} 0.992}     & {\color[HTML]{FF0000} 0.981}   \\ \hline
                                                                          & Pixel AUC                         & 0.988                            & 0.988                           & {\color[HTML]{FF0000} 0.992}      & 0.992                            & {\color[HTML]{0000FF} 0.990}   & 0.990                              & 0.994                             & 0.987                                & 0.983                          & 0.996                           & 0.964                          & 0.988                                & 0.961                                & 0.949                          & 0.990                            & {\color[HTML]{0000FF} 0.994}   \\
                                                                          & Pixel AP                          & 0.768                            & 0.653                           & 0.442                             & 0.627                            & 0.325                          & 0.537                              & 0.456                             & 0.870                                & 0.777                          & 0.354                           & 0.546                          & 0.372                                & 0.610                                & 0.477                          & 0.595                            & 0.561                          \\ 
\multirow{-3}{*}{PatchCore}                          & Pixel PRO                         & 0.957                            & 0.945                           & {\color[HTML]{0000FF} 0.958}      & 0.949                            & 0.939                          & 0.958                              & 0.974                             & 0.954                                & 0.945                          & 0.964                           & 0.906                          & 0.918                                & 0.906                                & 0.914                          & 0.961                            & 0.943                          \\ \hline
\end{tabular}}
\label{tab:fully-supervised}
\end{table*}

 \begin{figure*}[thbp]
    \centering
    \includegraphics[width=1\linewidth]{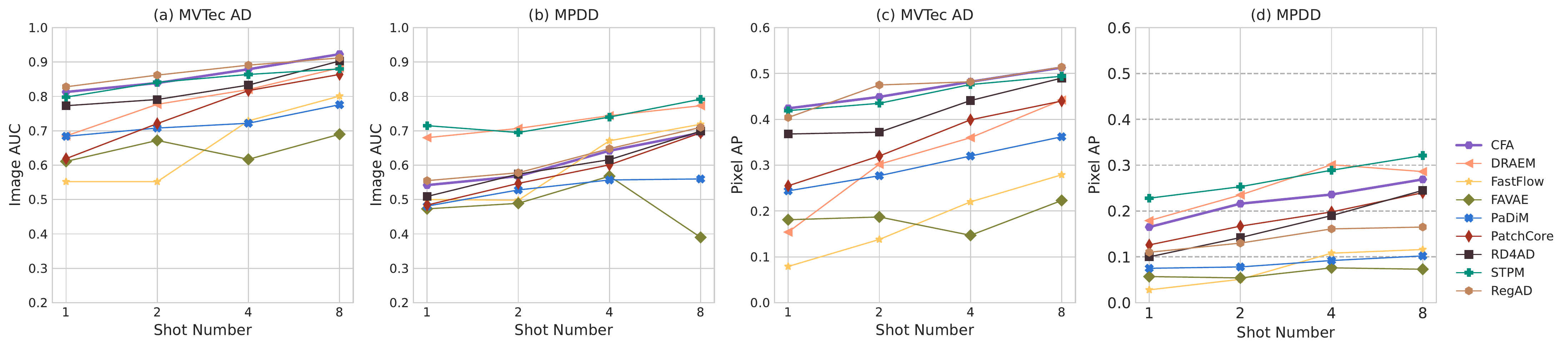}
    \caption{Few-shot IAD Benchmark on MVTec AD and MPDD. The Y-axis refers to the metric value and the X-axis denotes the shot number.}	
    \label{fig:fewshot_result}
\end{figure*}

\textbf{Discussions.} 
Fully supervised IAD algorithms use the distance between test and training samples to predict anomalies. The core idea is that the features of abnormal and normal samples are very different. For example, DevNet~\cite{pang2021explainable} proposes using the deviation loss function to enforce the statistical deviation of all anomalies from normal samples. From Table~\ref{tab:fully-supervised}, we can see that the performance of fully supervised IAD methods exceeds that of unsupervised methods (such as PatchCore) in pixel levels by a large margin. It justifies the effectiveness of incorporating an abnormal sample for training.

\textbf{Challenges.} 
 It is essential to improve the efficacy of the abnormal samples for the fully supervised IAD method. Because of the high cost associated with anomaly labels in real applications. On average, each worker spends three hours completing the pixel labels for each image. In subsequent research, researchers have to develop an improved fully supervised IAD algorithm with greater efficiency. The algorithm should reduce the number of abnormal samples for training while maintaining IAD accuracy. 

\subsection{Rotation Augmentation for Feature-Embedding based Few-Shot IAD}\label{sec:fewshot-rotation}

\textbf{Settings.} With respect to the training samples, we choose 1, 2, 4, and 8 to benchmark vanilla IAD methods. Details can be found in the few-shot setting of Sec.~\ref{sec:definition}-3. In addition, we make a comparison with RegAD~\cite{huang2022registration}, which is an advanced method of meta-learning. 

\textbf{Discussions.} 
Fig.~\ref{fig:fewshot_result} shows that CutPaste, STPM, and PatchCore's performance is comparable to the base model RegAD. Inspired by~\cite{Ni2020DataAF} and~\cite{Chen2019ImageBA}, we try to improve performance in IM few-shot settings using data augmentation. The statistical results of Table~\ref{tab:aug_fewshot_mvtec} indicate that most data augmentation methods are sufficient to improve the few-shot IAD performance. Further, we find that rotation is an optimal augmentation method because most of the real-time industrial image~\cite{Bergmann2019MVTecA, jezek2021deep} could be transformed into another image by rotation, such as metal\_nut and screw. 

\begin{table}[!h]
\Large
\renewcommand{\arraystretch}{1.2}
\centering
\caption{Image-level AUC-ROC $\uparrow$ on MVTec AD. The best and second-best results are marked in red and blue, respectively.}
\resizebox{\columnwidth}{!}{%
\begin{tabular}{l|l|lllllll}
\hline
 \rowcolor{blue!5} \textbf{Shot} &
 \textbf{Method} &
  \textbf{Vanilla} &
  \textbf{Rotation} &
  \textbf{Flip} &
  \textbf{Scale} &
  \textbf{Translate} &
  \textbf{ColorJitter} &
  \textbf{Perspective} \\ \hline
 &
  CFA &
  0.811 &
  {\color[HTML]{FF0000} 0.829} &
  0.811 &
  0.788 &
  0.802 &
  {\color[HTML]{0000FF} 0.814} &
  0.806 \\
 &
  CSFlow &
  0.708 &
  0.727 &
  0.708 &
  {\color[HTML]{0000FF} 0.742} &
  {\color[HTML]{FF0000} 0.750} &
  0.700 &
  0.713 \\
 &
  CutPaste &
  0.650 &
  0.701 &
  {\color[HTML]{0000FF} 0.702} &
  0.680 &
  0.679 &
  0.652 &
  {\color[HTML]{FF0000} 0.703} \\
 &
  DRAEM &
  0.683 &
  {\color[HTML]{0000FF} 0.718} &
  0.715 &
  0.714 &
  0.690 &
  0.687 &
  {\color[HTML]{FF0000} 0.741} \\
 &
  FastFlow &
  0.527 &
  0.618 &
  0.613 &
  {\color[HTML]{FF0000} 0.694} &
  {\color[HTML]{0000FF} 0.682} &
  0.578 &
  0.600 \\
 &
  FAVAE &
  0.651 &
  0.560 &
  {\color[HTML]{0000FF} 0.591} &
  0.600 &
  0.581 &
  0.588 &
  {\color[HTML]{FF0000} 0.626} \\
 &
  PaDiM &
  {\color[HTML]{0000FF} 0.684} &
  {\color[HTML]{FF0000} 0.697} &
  0.683 &
  0.669 &
  0.683 &
  0.681 &
  0.674 \\
 &
  PatchCore &
  0.788 &
  {\color[HTML]{FF0000} 0.805} &
  0.792 &
  0.788 &
  {\color[HTML]{0000FF} 0.800} &
  0.797 &
  0.789 \\
 &
  RD4AD &
  0.770 &
  0.805 &
  0.802 &
  0.799 &
  {\color[HTML]{FF0000} 0.823} &
  {\color[HTML]{0000FF} 0.816} &
  0.784 \\
 &
  SPADE &
  -- &
  -- &
  -- &
  -- &
  -- &
  -- &
  -- \\
\multirow{-11}{*}{1} &
  STPM &
  0.799 &
  {\color[HTML]{0000FF} 0.841} &
  0.814 &
  0.823 &
  {\color[HTML]{FF0000} 0.843} &
  0.831 &
  0.840 \\ \hline
 &
  CFA &
  0.839 &
  {\color[HTML]{FF0000} 0.860} &
  {\color[HTML]{0000FF} 0.853} &
  0.795 &
  0.825 &
  0.833 &
  0.843 \\
 &
  CSFlow &
  0.745 &
  {\color[HTML]{0000FF} 0.781} &
  0.773 &
  {\color[HTML]{FF0000} 0.783} &
  0.768 &
  0.754 &
  0.778 \\
 &
  CutPaste &
  0.697 &
  0.709 &
  {\color[HTML]{FF0000} 0.748} &
  0.659 &
  0.659 &
  0.611 &
  {\color[HTML]{0000FF} 0.726} \\
 &
  DRAEM &
  0.780 &
  0.765 &
  {\color[HTML]{0000FF} 0.774} &
  0.751 &
  0.771 &
  0.773 &
  {\color[HTML]{FF0000} 0.784} \\
 &
  FastFlow &
  0.555 &
  {\color[HTML]{0000FF} 0.743} &
  0.674 &
  {\color[HTML]{FF0000} 0.753} &
  0.731 &
  0.628 &
  0.617 \\
 &
  FAVAE &
  0.667 &
  {\color[HTML]{0000FF} 0.648} &
  {\color[HTML]{FF0000} 0.659} &
  0.595 &
  0.620 &
  0.642 &
  0.631 \\
 &
  PaDiM &
  0.708 &
  {\color[HTML]{FF0000} 0.734} &
  {\color[HTML]{0000FF} 0.731} &
  0.694 &
  0.716 &
  0.710 &
  0.714 \\
 &
  PatchCore &
  0.795 &
  {\color[HTML]{FF0000} 0.831} &
  0.805 &
  0.796 &
  {\color[HTML]{0000FF} 0.806} &
  0.793 &
  0.802 \\
 &
  RD4AD &
  0.798 &
  {\color[HTML]{0000FF} 0.832} &
  0.816 &
  {\color[HTML]{FF0000} 0.835} &
  0.817 &
  {\color[HTML]{FF0000} 0.835} &
  0.820 \\
 &
  SPADE &
  -- &
  0.737 &
  0.731 &
  0.710 &
  0.734 &
  0.733 &
  0.744 \\
\multirow{-11}{*}{2} &
  STPM &
  0.840 &
  0.839 &
  {\color[HTML]{0000FF} 0.850} &
  0.848 &
  {\color[HTML]{FF0000} 0.851} &
  0.849 &
  0.838 \\ \hline
 &
  CFA &
  0.879 &
  {\color[HTML]{0000FF} 0.891} &
  0.861 &
  0.818 &
  0.860 &
  {\color[HTML]{FF0000} 0.895} &
  0.883 \\
 &
  CSFlow &
  0.785 &
  {\color[HTML]{FF0000} 0.841} &
  0.815 &
  {\color[HTML]{0000FF} 0.825} &
  0.810 &
  0.806 &
  0.738 \\
 &
  CutPaste &
  0.728 &
  {\color[HTML]{FF0000} 0.771} &
  {\color[HTML]{0000FF} 0.714} &
  0.519 &
  0.674 &
  0.621 &
  0.671 \\
 &
  DRAEM &
  0.820 &
  0.798 &
  0.802 &
  0.819 &
  {\color[HTML]{0000FF} 0.824} &
  0.823 &
  {\color[HTML]{FF0000} 0.826} \\
 &
  FastFlow &
  0.693 &
  0.741 &
  0.745 &
  {\color[HTML]{FF0000} 0.790} &
  {\color[HTML]{0000FF} 0.782} &
  0.723 &
  0.734 \\
 &
  FAVAE &
  0.655 &
  0.626 &
  {\color[HTML]{0000FF} 0.669} &
  0.605 &
  0.639 &
  {\color[HTML]{FF0000} 0.670} &
  {\color[HTML]{FF0000} 0.623} \\
 &
  PaDiM &
  0.722 &
  {\color[HTML]{FF0000} 0.734} &
  0.723 &
  0.689 &
  0.720 &
  {\color[HTML]{0000FF} 0.727} &
  0.716 \\
 &
  PatchCore &
  0.844 &
  {\color[HTML]{FF0000} 0.872} &
  0.826 &
  0.844 &
  0.848 &
  {\color[HTML]{0000FF} 0.852} &
  0.847 \\
 &
  RD4AD &
  0.838 &
  {\color[HTML]{FF0000} 0.897} &
  0.871 &
  0.866 &
  {\color[HTML]{0000FF} 0.876} &
  0.872 &
  0.861 \\
 &
  SPADE &
  -- &
  {\color[HTML]{FF0000} 0.764} &
  0.739 &
  0.749 &
  0.758 &
  0.757 &
  {\color[HTML]{0000FF} 0.759} \\
\multirow{-11}{*}{4} &
  STPM &
  0.864 &
  0.869 &
  0.865 &
  {\color[HTML]{0000FF} 0.876} &
  0.875 &
  0.868 &
  {\color[HTML]{FF0000} 0.885} \\ \hline
 &
  CFA &
  0.923 &
  0.913 &
  0.888 &
  0.875 &
  {\color[HTML]{0000FF} 0.920} &
  0.919 &
  {\color[HTML]{FF0000} 0.924} \\
 &
  CSFlow &
  0.856 &
  {\color[HTML]{FF0000} 0.900} &
  0.863 &
  {\color[HTML]{FF0000} 0.900} &
  {\color[HTML]{0000FF} 0.893} &
  0.892 &
  0.861 \\
 &
  CutPaste &
  0.705 &
  {\color[HTML]{0000FF} 0.808} &
  {\color[HTML]{FF0000} 0.857} &
  0.473 &
  0.694 &
  0.607 &
  0.673 \\
 &
  DRAEM &
  0.872 &
  0.892 &
  {\color[HTML]{0000FF} 0.905} &
  0.892 &
  0.900 &
  0.904 &
  {\color[HTML]{FF0000} 0.919} \\
 &
  FastFlow &
  0.828 &
  0.809 &
  0.793 &
  {\color[HTML]{FF0000} 0.843} &
  {\color[HTML]{0000FF} 0.826} &
  0.809 &
  0.810 \\
 &
  FAVAE &
  0.701 &
  0.651 &
  {\color[HTML]{0000FF} 0.670} &
  0.619 &
  0.611 &
  {\color[HTML]{FF0000} 0.680} &
  0.668 \\
 &
  PaDiM &
  0.776 &
  {\color[HTML]{FF0000} 0.810} &
  0.739 &
  0.734 &
  0.761 &
  {\color[HTML]{0000FF} 0.779} &
  0.771 \\
 &
  PatchCore &
  0.891 &
  {\color[HTML]{FF0000} 0.916} &
  0.876 &
  0.889 &
  0.898 &
  0.883 &
  {\color[HTML]{0000FF} 0.891} \\
 &
  RD4AD &
  0.903 &
  {\color[HTML]{FF0000} 0.933} &
  0.911 &
  0.917 &
  {\color[HTML]{0000FF} 0.921} &
  0.912 &
  0.916 \\
 &
  SPADE &
  0.781 &
  {\color[HTML]{FF0000} 0.794} &
  0.770 &
  0.783 &
  {\color[HTML]{0000FF} 0.791} &
  0.788 &
  0.787 \\
\multirow{-11}{*}{8} &
  STPM &
  0.865 &
  0.896 &
  0.870 &
  0.910 &
  {\color[HTML]{FF0000} 0.914} &
  {\color[HTML]{0000FF} 0.912} &
  0.904 \\ \hline
\end{tabular}%
}
\label{tab:aug_fewshot_mvtec}
\end{table}

\textbf{Challenges.} 
 Synthesis of abnormal samples is essential but difficult. Previously, researchers focused on developing data augmentation methods for normal images. However, little effort is made to synthesize abnormal samples. Because of the fault-free production line in industrial environments, collecting large quantities of abnormal samples is very difficult. In the future, more attention should be paid to abnormal synthesizing methods, such as CutPaste~\cite{li2021cutpaste} and DRAEM~\cite{zavrtanik2021draem}.

\subsection{Importance Re-weighting for Noisy IAD}\label{sec:angle_noisy}

\textbf{Settings.} 
 Using the noise setting introduced in Sec.~\ref{sec:definition}-4, we create a noisy label training dataset by injecting a variety of abnormal samples from the original test set. Specifically, we borrowed the settings from SoftPatch~\cite{jiang2022softpatch}. The abnormal sample accounts for mainly 20\% of all training data (called noise ratio). The noise ratio is 5\% to 20\% and its step size is 5\%. Due to the limited size of the test category, we selected a training sample of up to 75\% of the abnormal sample of the test dataset. At the training stage, the labels observed in all training datasets are normal. During the test phase, abnormal samples are no longer evaluated. In the noise settings, we benchmark the vanilla and noisy IAD baseline method, IGD~\cite{Chen2022DeepOC}.

\begin{table}[th]
\centering
\caption{Image AUC $\uparrow$ of Noisy IAD methods on MVTec AD and MPDD under various noise ratios. The best and second-best results are marked in red and blue, respectively.}
\renewcommand{\arraystretch}{1.2}
\resizebox{1\linewidth}{!}{
\begin{tabular}{l|llll|llll}
\hline
\rowcolor{blue!5}  \textbf{Dataset}        & \multicolumn{4}{c|}{\textbf{MVTec AD}}                                                                                             & \multicolumn{4}{c}{\textbf{MPDD}}     \\ \hline
\rowcolor{blue!5} \textbf{Method}   & \textbf{0.05}                         & \textbf{0.1}                          & \textbf{0.15}                         & \textbf{0.2}                          & \textbf{0.05}                         & \textbf{0.1}                          & \textbf{0.15 }                        & \textbf{0.2 }                         \\ \hline
CFA       & 0.969                        & 0.962                        & 0.949                        & 0.946                        & {\color[HTML]{0000FF} 0.894} & 0.808                        & 0.839                        & 0.782                        \\
CS-Flow   & 0.939                        & 0.894                        & 0.904                        & 0.874                        & 0.773                        & 0.789                        & {\color[HTML]{FE0000} 0.861} & 0.727                        \\
CutPaste  & 0.864                        & 0.900                        & 0.851                        & 0.907                        & 0.773                        & 0.671                        & 0.653                        & 0.683                        \\
DRAEM     & 0.954                        & 0.912                        & 0.869                        & 0.897                        & 0.810                        & 0.760                        & 0.726                        & 0.693                        \\
FastFlow  & 0.885                        & 0.867                        & 0.854                        & 0.855                        & 0.836                        & 0.725                        & 0.725                        & 0.691                        \\
FAVAE     & 0.768                        & 0.801                        & 0.798                        & 0.814                        & 0.536                        & 0.476                        & 0.482                        & 0.525                        \\
IGD       & 0.805                        & 0.801                        & 0.782                        & 0.790                        & 0.797                        & 0.786                        & 0.789                        & 0.756                        \\
PaDiM     & 0.890                        & 0.907                        & 0.899                        & 0.906                        & 0.675                        & 0.580                        & 0.626                        & 0.608                        \\
PatchCore & {\color[HTML]{FE0000} 0.990} & {\color[HTML]{FE0000} 0.986} & {\color[HTML]{FE0000} 0.975} & {\color[HTML]{0000FF} 0.980} & 0.857                        & 0.790                        & 0.782                        & 0.763                        \\
RD4AD     & {\color[HTML]{0000FF} 0.989} & {\color[HTML]{0000FF} 0.984} & {\color[HTML]{FE0000} 0.975} & {\color[HTML]{FE0000} 0.983} & {\color[HTML]{FE0000} 0.909} & {\color[HTML]{FE0000} 0.837} & 0.826                        & {\color[HTML]{FE0000} 0.824} \\
SPADE     & 0.854                        & 0.861                        & 0.855                        & 0.859                        & 0.784                        & 0.728                        & 0.737                        & 0.719                        \\
STPM      & 0.909                        & 0.877                        & 0.892                        & 0.915                        & 0.862                        & {\color[HTML]{0000FF} 0.830} & {\color[HTML]{0000FF} 0.848} & {\color[HTML]{0000FF} 0.809} \\ \hline
\end{tabular}}
\label{tab:noisy_comparsion}
\end{table}

\textbf{Discussions.} 
Based on the statistical findings presented in Table~\ref{tab:noisy_comparsion}, it has been discovered that feature embedding-based methods are more effective than IGD when dealing with limited noise levels ($\leq$ 0.15). In order to identify the cause, we have conducted a thorough investigation into the representative feature-embedding based method, PatchCore~\cite{roth2022towards}.
During the training phase, the objective is to establish a memory bank that stores neighborhood-aware features from all normal samples. The algorithm used to construct the feature memory is detailed in Algorithm~\ref{alg:aug}.

\begin{algorithm}[ht]
    \small
    \caption{PatchCore Pipeline}\label{alg:aug}
    \SetKwInOut{KwIn}{Input}
    \SetKwInOut{KwOut}{Output}
    \KwIn{ImageNet pretrained $\phi$, all normal samples $\mathcal{X}_{N}$ patch feature extractor $\mathcal{P}$, memory size target $l$, random linear projection $\psi$.}
    \KwOut{Patch-level augmented memory bank $\mathcal{M}$.}
    $\mathcal{M} \leftarrow \left\{ \right\}$\;
    \For{$x_{i} \in \mathcal{X_{N}}$}
    {        
        $\mathcal{M} \leftarrow \mathcal{M} \cup \mathcal{P}(\phi(x_{i}))$\;
    }
    $\mathcal{M}_{C} \leftarrow \left\{ \right\}$ \textcolor{gray}{// Apply coreset sampling for memory bank}\\
    \For{$i \in \left [0,\cdots, l-1  \right ]$}
    {
        $m_{i} \leftarrow \underset{m \in \mathcal{M} - \mathcal{M_{C}}}{\argmax} \underset{n \in \mathcal{M_{C}}} {\min} \left\| \psi(m) - \psi(n) \right\|_{2}$\;
        $\mathcal{M_{C}} \leftarrow \mathcal{M_{C}} \cup \left\{m_{i} \right\}$\;
    }
    $\mathcal{M} \leftarrow \mathcal{M_{C}}$.
\end{algorithm}

\begin{table*}[htb]
\centering
\caption{Performance in terms of FM on the continual setting. The best and second-best results are marked in red and blue, respectively.}
\renewcommand{\arraystretch}{1.2}
\resizebox{\linewidth}{!}{
\begin{tabular}{l|lllll|lllll}
\hline
 \rowcolor{blue!5} \textbf{Dataset} & \multicolumn{5}{c|}{ { \textbf{ MVTec AD}}}                                                                   & \multicolumn{5}{c}{ {  \textbf{MPDD}}}     \\ \hline
  \rowcolor{blue!5}{  }                   & \multicolumn{2}{c|}{\textbf{Image}}                                     & \multicolumn{3}{c|}{\textbf{Pixel}}                                                    & \multicolumn{2}{c|}{\textbf{Image}}                                     & \multicolumn{3}{c}{\textbf{Pixel}}                                                    \\
 \rowcolor{blue!5} \textbf{Method} & \multicolumn{1}{c}{\textbf{AUC$\uparrow$ / FM$\downarrow$}} & \multicolumn{1}{c|}{\textbf{AP$\uparrow$ / FM$\downarrow$}}            & \multicolumn{1}{c}{\textbf{AUC$\uparrow$ / FM$\downarrow$}} & \multicolumn{1}{c}{\textbf{AP$\uparrow$ / FM$\downarrow$}} & \multicolumn{1}{c|}{\textbf{PRO$\uparrow$ / FM$\downarrow$}} & \multicolumn{1}{c}{\textbf{AUC$\uparrow$ / FM$\downarrow$}} & \multicolumn{1}{c|}{\textbf{AP$\uparrow$ / FM$\downarrow$}}            & \multicolumn{1}{c}{\textbf{AUC$\uparrow$ / FM$\downarrow$}} & \multicolumn{1}{c}{\textbf{AP$\uparrow$ / FM$\downarrow$}} & \multicolumn{1}{c}{\textbf{PRO$\uparrow$ / FM$\downarrow$}} \\ \hline
PatchCore                                                         & {\color[HTML]{FE0000}0.919} / {\color[HTML]{FE0000}0.003}             & \multicolumn{1}{c|}{{\color[HTML]{FE0000}0.971} / {\color[HTML]{FE0000}0.001}} & {\color[HTML]{FE0000}0.944} / {\color[HTML]{FE0000}0.002}             & {\color[HTML]{FE0000}0.481} / {\color[HTML]{FE0000}0.005}          & {\color[HTML]{FE0000}0.847} / {\color[HTML]{FE0000}0.001}            & {\color[HTML]{FE0000}0.786} / {\color[HTML]{FE0000}0.015}             & \multicolumn{1}{c|}{{\color[HTML]{FE0000}0.851} / {\color[HTML]{FE0000}0.021}} & {\color[HTML]{FE0000}0.963} / {\color[HTML]{FE0000}0}                 & {\color[HTML]{FE0000}0.358} / {\color[HTML]{FE0000}0.0004}         & {\color[HTML]{FE0000}0.838} / {\color[HTML]{FE0000}0.0001}               \\
CFA                                                               & {\color[HTML]{0000FF}0.623} / 0.361             & \multicolumn{1}{c|}{{\color[HTML]{0000FF}0.813} / 0.181} & {\color[HTML]{0000FF}0.753} / 0.217             & {\color[HTML]{0000FF}0.176} / 0.359          & 0.535 / 0.363            & 0.506 / 0.415             & \multicolumn{1}{c|}{0.609 / 0.307} & 0.818 / 0.131             & 0.116 / 0.164          & 0.557 / 0.300           \\
SPADE                                                             & 0.571 / 0.284             & \multicolumn{1}{c|}{0.781 / 0.159} & 0.746 / 0.208             & 0.151 / 0.319          & {\color[HTML]{0000FF}0.570} / 0.324            & 0.412 / 0.396             & \multicolumn{1}{c|}{0.576 / 0.259}  & {\color[HTML]{0000FF}0.912} / {\color[HTML]{0000FF}0.069}             & 0.139 / 0.202          & {\color[HTML]{0000FF}0.732} / {\color[HTML]{0000FF}0.193}           \\
PaDiM                                                             &      0.545 / 0.368                    & \multicolumn{1}{c|}{0.767 / 0.192}              &             0.697 / 0.269              &   0.086 / 0.366                     &          0.441 / 0.472                &        0.461 / 0.262                   & \multicolumn{1}{c|}{0.593 / 0.195}              &         0.841 / 0.116                  &        0.053 / 0.106                &     0.529 / 0.332                    \\
RD4AD                                                             & 0.596 / 0.392             & \multicolumn{1}{c|}{0.800 / 0.194} & {\color[HTML]{0000FF}0.753} / 0.222             & 0.143 / 0.425          & 0.531 / 0.401            & 0.646 / 0.344             & \multicolumn{1}{c|}{{\color[HTML]{0000FF}0.833} / 0.162} & 0.684 / 0.300             & {\color[HTML]{0000FF}0.158} / 0.407          & 0.532 / 0.414           \\
STPM                                                              & 0.576 / 0.324             & \multicolumn{1}{c|}{0.786 / 0.163} & 0.625 / 0.277             & 0.109 / 0.352          & 0.322 / 0.446            & 0.499 / 0.339             & \multicolumn{1}{c|}{0.617 / 0.267} & 0.328 / 0.648             & 0.099 / 0.217          & 0.217 / 0.701           \\
CutPaste                                                          & 0.312 / 0.509             & \multicolumn{1}{c|}{0.650 / 0.270} &  --                         & --                       &     --                     & {\color[HTML]{0000FF}0.665} / {\color[HTML]{0000FF}0.045}             & \multicolumn{1}{c|}{0.724 / {\color[HTML]{0000FF}0.061}} &   --                        &       --                 &   --                      \\
CSFlow                                                          & 0.538 / 0.426               & \multicolumn{1}{c|}{ 0.762 / 0.224 }& --                           & --                        & --                         &0.632 / 0.343 &\multicolumn{1}{c|}{ 0.692 / 0.290} & --              & --   &  --  \\
FastFlow                                                          & 0.512 / 0.279             & \multicolumn{1}{c|}{0.713 / 0.154} & 0.519 / 0.380             & 0.004 / 0.214          & 0.152 / 0.562            & 0.542 / 0.272             & \multicolumn{1}{c|}{0.643 / 0.173} & 0.269 / 0.506             & 0.015 / {\color[HTML]{0000FF}0.071}          & 0.061 / 0.448           \\
FAVAE                                                             & 0.547 / {\color[HTML]{0000FF}0.101}             & \multicolumn{1}{c|}{0.772 / {\color[HTML]{0000FF}0.055}} & 0.673 / {\color[HTML]{0000FF}0.107}             & 0.082 / {\color[HTML]{0000FF}0.082}          & 0.390 / {\color[HTML]{0000FF}0.157}            & 0.581 / 0.183             & \multicolumn{1}{c|}{0.792 / 0.099} & 0.636 / 0.197             & 0.098 / 0.168          & 0.365 / 0.267           \\ \hline
\rowcolor{black!5}DNE                                                               & 0.537 / 0.299             & \multicolumn{1}{c|}{0.533 / 0.293} &       --                    &         --               &          --                & 0.413 / 0.394             & \multicolumn{1}{c|}{0.362 / 0.428} &               --            &         --               &              --           \\ \hline
\end{tabular}
}
\label{tab:continual_comparsion}
\end{table*}

\begin{table}[th]
\centering
\caption{Image AUROC $\uparrow$ of PatchCore under different noise ratios and neighbourhood numbers. The best and second-best results are marked in red and blue, respectively.}
\renewcommand{\arraystretch}{1.2}
\resizebox{0.95\linewidth}{!}{
\begin{tabular}{l|lllll}
\hline
\rowcolor{blue!5} & \multicolumn{5}{c}{ {\textbf{ Neighbour Number}}} \\\cline{2-6} 
\rowcolor{blue!5} \textbf{Noise Ratio} & \textbf{1 (No Re-weighting)} & \textbf{2}    & \textbf{3}       & \textbf{6}    & \textbf{9}      \\ \hline
0.05                                                     & {\color[HTML]{FE0000} 0.990}  & {\color[HTML]{0000FF} 0.988}                        & {\color[HTML]{FE0000} 0.990}  & 0.987                        & {\color[HTML]{FE0000} 0.990}  \\
0.1                                                      & {\color[HTML]{0000FF} 0.986} & {\color[HTML]{FE0000} 0.989} & {\color[HTML]{0000FF} 0.986} & 0.984                        & 0.983                        \\
0.15                                                     & 0.975                        & 0.983                        & 0.979                        & {\color[HTML]{FE0000} 0.986} & {\color[HTML]{0000FF} 0.984} \\
0.2                                                      & 0.980                         & {\color[HTML]{0000FF} 0.982} & 0.976                        & {\color[HTML]{FE0000} 0.986} & {\color[HTML]{0000FF} 0.982} \\ \hline
\end{tabular}}
\label{tab:reweighting-patchcore}
\end{table}

\begin{table*}[htb]
\caption{Benchmark of the representative vanilla IAD algorithms in IM-IAD setting, including few-shot, noisy label, continual learning. The best and second-best results are marked in red and blue, respectively.}
\renewcommand{\arraystretch}{1}
\resizebox{\linewidth}{!}{
\begin{tabular}{ll|lllll|lllll}
\hline
\rowcolor{blue!5}  \textbf{Dataset}         &           & \multicolumn{5}{c|}{\textbf{MVTec AD}}                               & \multicolumn{5}{c}{\textbf{MPDD}}                                   \\ \hline
\rowcolor{blue!5} \textbf{Metric}                     & \textbf{Method}   & \textbf{Vanilla} & \textbf{Noisy-0.15} & \textbf{Few-shot-8} & \textbf{Continual} & \textbf{Mean} & \textbf{Vanilla} & \textbf{Noisy-0.15} & \textbf{Few-shot-8} & \textbf{Continual} &  \textbf{Mean} \\ \hline

\multirow{11}{*}{Image AUC $\uparrow$} & CFA       & 0.981   & {\color[HTML]{0000FF} 0.949}            & {\color[HTML]{FE0000} 0.923}             & {\color[HTML]{0000FF} 0.623}  & {\color[HTML]{0000FF} 0.869}   & 0.923   & 0.839            & 0.694             & 0.506  & 0.740  \\
                            & CSFlow    & 0.952   & 0.904            & 0.845             & 0.539  & 0.810   & {\color[HTML]{FE0000} 0.973}   & {\color[HTML]{FE0000} 0.861}            & 0.741             & 0.633   & {\color[HTML]{0000FF} 0.802}  \\
                            & CutPaste  & 0.918   & 0.851            & 0.705             & 0.312    & 0.696 & 0.771 & 0.653            & 0.577             & {\color[HTML]{0000FF} 0.665}   & 0.667  \\
                            & DRAEM     & 0.981   & 0.869            & 0.883             & 0.595 & 0.832    & 0.941   & 0.726            & {\color[HTML]{0000FF} 0.773}             & 0.440  & 0.720   \\
                            & FastFlow  & 0.905   & 0.854            & 0.801             & 0.512   & 0.768  & 0.887   & 0.725            & 0.719             & 0.543  & 0.718   \\
                            & FAVAE     & 0.793   & 0.798            & 0.690             & 0.547  & 0.707   & 0.570   & 0.482            & 0.390             & 0.582  & 0.506   \\
                            & PaDiM     & 0.908   & 0.899            & 0.776             & 0.545  & 0.782   & 0.706   & 0.626            & 0.560             & 0.461   & 0.588  \\
                            & PatchCore & {\color[HTML]{FE0000} 0.992}   & {\color[HTML]{FE0000} 0.975}            & 0.864             & {\color[HTML]{FE0000} 0.919}   & {\color[HTML]{FE0000} 0.938}  & {\color[HTML]{0000FF} 0.948}   & 0.782            & 0.694             & {\color[HTML]{FE0000} 0.787}  & {\color[HTML]{FE0000} 0.803}  \\
                            & RD4AD     & {\color[HTML]{0000FF} 0.986}   & {\color[HTML]{FE0000} 0.975}            & {\color[HTML]{0000FF} 0.903}             & 0.596 & 0.865     & 0.927   & 0.826            & 0.699             & 0.646   & 0.775  \\
                            & SPADE     & 0.854   & 0.855            & 0.781             & 0.571   & 0.765  & 0.784   & 0.737            & 0.594             & 0.412   & 0.632  \\
                            & STPM      & 0.924   & 0.892            & 0.880             & 0.576  & 0.818   & 0.876   & {\color[HTML]{0000FF} 0.848}            & {\color[HTML]{FE0000} 0.792}             & 0.499 & 0.754     \\ \hline
\multirow{11}{*}{Pixel AP $\uparrow$}  & CFA       & 0.538   & 0.394            & {\color[HTML]{FE0000} 0.513}             & {\color[HTML]{0000FF} 0.177}  & 0.406   & 0.283   & 0.212            & 0.269             & 0.117   & 0.220  \\
                            & CSFlow    & --       & --                & --                 & --         & --       & --                & --                 & --     &  -- & --    \\
                            & CutPaste  & --       & --                & --                 & --         & --       & --                & --                 & --     &  -- & --    \\
                            & DRAEM     & {\color[HTML]{FE0000} 0.689}   & 0.337            & 0.442             & 0.098   & 0.391  & 0.288   & 0.174            & {\color[HTML]{0000FF} 0.286}             & 0.131  & 0.220   \\
                            & FastFlow  & 0.398   & 0.313            & 0.279             & 0.044  & 0.259    & 0.115   & 0.055            & 0.116             & 0.015  & 0.075   \\
                            & FAVAE     & 0.307   & 0.319            & 0.223             & 0.083  & 0.233   & 0.088   & 0.080            & 0.073             & 0.099    & 0.085 \\
                            & PaDiM     & 0.452   & 0.454            & 0.362             & 0.086   & 0.339  & 0.155   & 0.136            & 0.102             & 0.053   & 0.112  \\
                            & PatchCore & 0.561   & {\color[HTML]{0000FF} 0.500}            & 0.440             & {\color[HTML]{FE0000} 0.481} & {\color[HTML]{FE0000} 0.496}    & {\color[HTML]{0000FF} 0.432}   & 0.254            & 0.240             & {\color[HTML]{FE0000} 0.358}   & {\color[HTML]{FE0000} 0.321} \\
                            & RD4AD     & {\color[HTML]{0000FF} 0.580}   & {\color[HTML]{FE0000} 0.532}            & 0.490             & 0.143  & {\color[HTML]{0000FF} 0.436}   & {\color[HTML]{FE0000} 0.455}   & {\color[HTML]{FE0000} 0.374}            & 0.245             & {\color[HTML]{0000FF} 0.158}   & {\color[HTML]{0000FF} 0.308}  \\
                            & SPADE     & 0.471   & 0.410            & 0.405             & 0.151   & 0.359  & 0.342   & {\color[HTML]{0000FF} 0.309}            & 0.261             & 0.140  & 0.263    \\
                            & STPM      & 0.518   & 0.465            & {\color[HTML]{0000FF} 0.494}             & 0.110  & 0.354   & 0.354   & {\color[HTML]{0000FF} 0.309}            & {\color[HTML]{FE0000} 0.321}             & 0.100 & 0.271 \\ \hline
\end{tabular}}
\label{tab:uniform_setting}
\end{table*}

By default, WideResNet-50~\cite{zagoruyko2016wide} is set as a feature extraction model. Coreset sampling~\cite{Sener2018ActiveLF} for memory banks aims to balance the size of memory banks with IAD performance. In inference, the test image will be predicted as an anomaly if at least one patch is anomalous, and pixel-level anomaly segmentation will be computed via the score of each patch feature. In particular, with the normal patch feature bank $\mathcal{M}$, the image-level anomaly score $s$ for the test image $x^{test}$ is computed by the maximum score $s^{*}$ between the test image's patch feature $\mathcal{P}(x^{test})$ and its nearest neighbor $m^{*}$ in $\mathcal{M}$: 
\begin{equation}\label{eq:find_neighbour}
    m^{*} = \underset{m^{test} \in \mathcal{P}(x^{test})}{\argmax} \underset{m \in \mathcal{M}}{\argmin}\left\| m^{test} - m\right\|_{2},
\end{equation}
\begin{equation}\label{eq:anomaly_score}
    s^{*} = \left\| m^{test} - m^{*}\right\|_{2}.
\end{equation}
To enhance model robustness, PatchCore employs the importance re-weighting~\cite{Liu2014ClassificationWN} to tune the anomaly score $s$.
\begin{equation}
    s = \left ( 1- \frac{\mathrm{exp}\left\|m^{test, *} - m^{*} \right\|_{2}}{\sum_{m \in \mathcal{N}_{b}(m^{*})}\mathrm{exp} \left\|m^{test, *} - m^{*} \right\|_{2}} \right ) \cdot s^{*},
\end{equation}
where $\mathcal{N}_{b}(m^{*})$ denotes $b$ nearest patch-features in $\mathcal{M}$ for test patch-feature $m^{*}$. Furthermore, we conducted ablation studies to verify the effect of re-weighting. The statistical result of Table~\ref{tab:reweighting-patchcore} shows that re-weighting improves the resilience of IAD algorithms when the noise ratio is larger than 10\%.

\textbf{Challenges.} 
 The selection of samples is essential to improve IAD's robustness. Sample selection has excellent potential in the future, as it can be seamlessly integrated with current IAD methods. For example, Song \textit{et al.}~\cite{Song2019SELFIERU} create a discriminative network to distinguish true-labeled cases from noisy training data, \textit{i.e.}, one for sample selection and the other for anomalous detection.

\subsection{Memory Bank-based Methods for Continual IAD}

\textbf{Settings.} We benchmark the vanilla methods and the continual baseline IAD method, DNE~\cite{li2022towards} according to the continual setting in Sec.~\ref{sec:definition}-5.

\textbf{Discussions.} 
Table~\ref{tab:continual_comparsion} indicates that memory bank-based methods provide satisfactory performance to overcome the catastrophic forgetting issue among all unsupervised IAD, even though these methods are not explicitly proposed for continual IAD settings. The main idea is that the replay method is perfectly appropriate for memory-based IAD algorithms. Memory bank-based methods can easily add new memory to reduce forgetting when learning new tasks. Ideally, the memory bank should work for the test mechanism. During the test, the memory bank-based method can easily combine the nearest-mean classifier to locate the nearest memory when the test sample is provided. 
 
\textbf{Challenges.}
In the continual IAD setting, limiting the size of the memory database and minimizing interference from previous tasks is of utmost importance. When learning new tasks, simply adding more memory significantly slows down the inference time and increases storage usage. For a fixed storage need, we suggest using the storage sampling~\cite{Rolnick2019ExperienceRF} to limit the number of instances that can be stored. Furthermore, the memory of the new task may overlap with the memory of the previous task, causing IAD performance to decrease. Consequently, the main challenge is to design a new constraint optimization loss function to avoid task interference.

\subsection{Uniform View on IM-IAD}
\textbf{Settings.} We benchmark vanilla algorithms in a uniform setting, including few-shot, noisy, and continual settings, which are described in Sec.~\ref{sec:definition}.

\textbf{Discussions.} 
Table~\ref{tab:uniform_setting} shows that PatchCore achieves the SOTA result in IM-IAD. However, as shown in Fig.~\ref{fig:efficiency}, PatchCore is inferior to other IAD algorithms regarding GPU memory usage and inference time, making its use challenging in real scenarios. Furthermore, maintaining the size of the memory bank is a critical issue for realistic deployment if the number of object classes exceeds 100. Therefore, to bridge the gap between academic research and industry, there are still opportunities for IAD algorithms.

\textbf{Challenges.}
Multi-objective neural architecture search (NAS) is a considerably promising direction for IAD in finding the optimal trade-off architecture. If the IAD model can be applied in practice, many objectives need to be satisfied, like fast speed of inference, low memory usage, and continual learning ability. Because the memory capacity of edge devices used in production lines is heavily restricted. Moreover, most IAD methods cannot predict the arrival of new test samples instantly when test samples are sequentially streamed on the product line.

\section{Conclusion} \label{sec:conclusion}
We introduce IM-IAD in this paper, thus far the most complete industrial manufacturing-based IAD benchmark with 19 algorithms, 7 datasets, and 5 settings. We have gained insights into the importance of partial labels for fully supervised learning, the influence of noise ratio in noisy settings, the design principles for continual learning, and the important role of global features for logical anomalies. On top of these, we present several intriguing future lines for industrial IAD.

\section*{Acknowledgments} 
This work is supported by the National Natural Science Foundation of China (Grant NO. 62122035 and 62206122).

\bibliographystyle{ieee_fullname}
\bibliography{reference}

\begin{IEEEbiography}[{\includegraphics[width=1in,height=1.25in,clip,keepaspectratio]{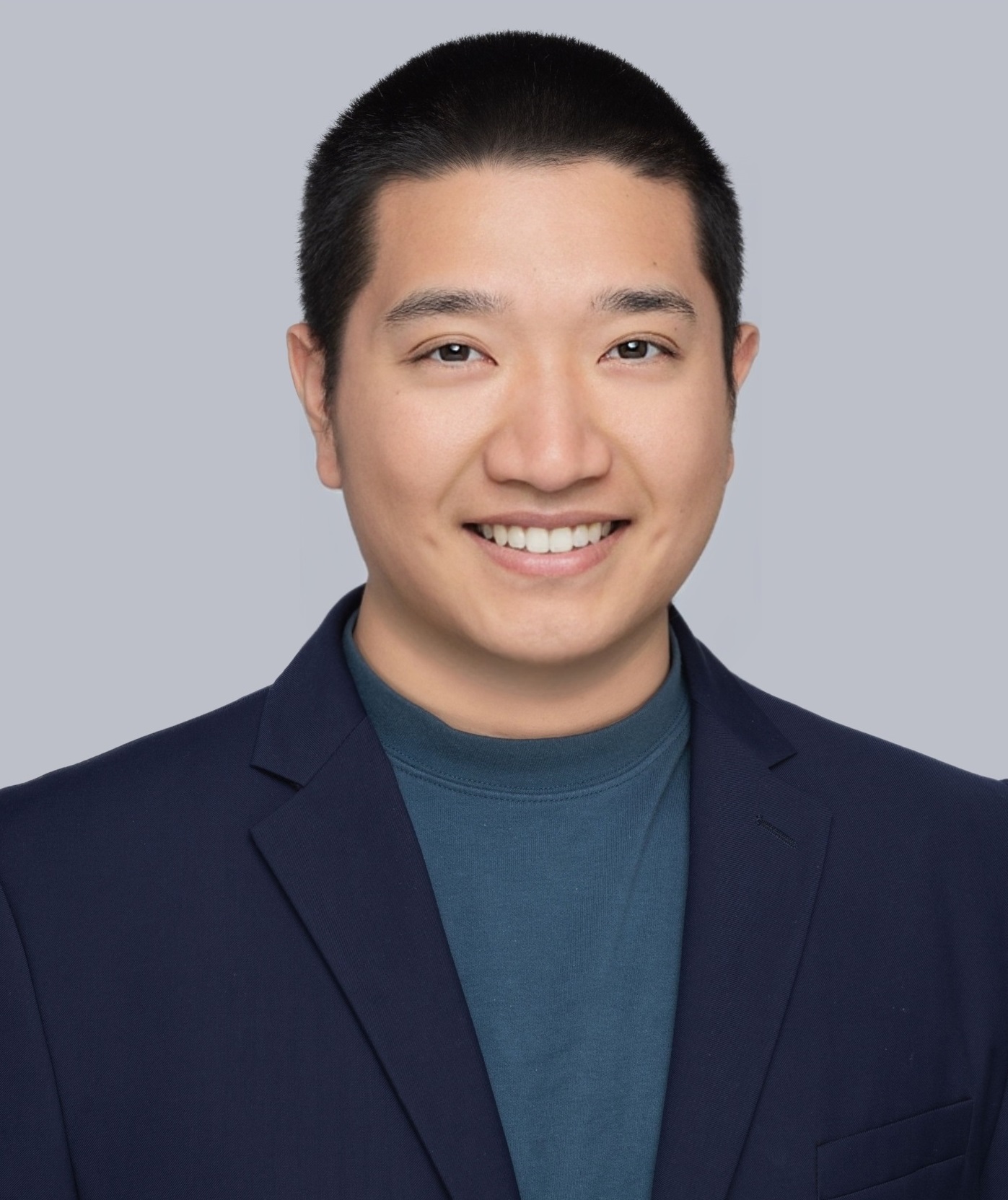}}]{Guoyang Xie} received the B.Sc. degree  from the University of Electronic Science and Technology of China in 2013,  the MPhil. degree from the Hong Kong University of Science and Technology in in 2015 and the Ph.D. degree at the University of Surrey, Guildford, United Kingdom in 2023. He is presently an Algorithm Manager in CATL, Ningde, China. Prior to that, he was the Principle Perception Algorithm Engineer in Baidu and GAC, respectively.  His research interests include AI for manufacturing and medical imaging, industrial image anomaly detection, robot learning and data synthesis. He has published more than 14 papers in peer-reviewed top-tier conferences (NeurIPS, ICLR, ACM MM, CVPR) and journals (ACM Computing Survey, TCYB, Neurocomputing, Complex \& Intelligent System). He has also served as the reviewer for ICML, NeurIPS, ICLR, AAAI, ACM MM, TEVC, TNNLS, TSMC, TETCI.
\end{IEEEbiography}

\begin{IEEEbiography}[{\includegraphics[width=1in,height=1.25in,clip,keepaspectratio]{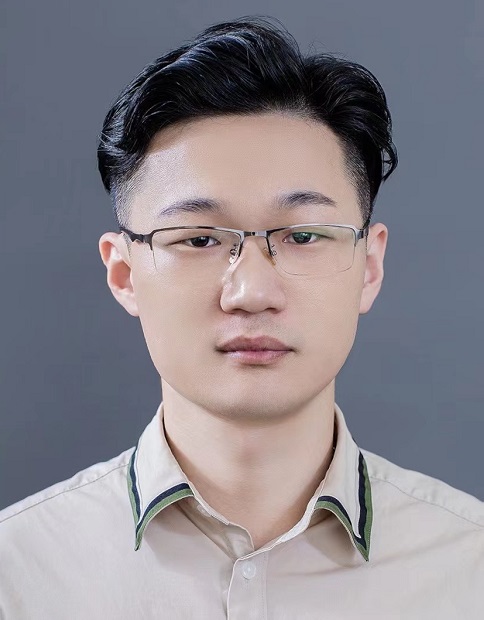}}]{Jinbao Wang} was born in Hebei on 12 Nov. 1989 and received a PhD degree from the University of Chinese Academy of Sciences (UCAS) in 2019. He currently is an Assistant Professor at the National Engineering Laboratory for Big Data System Computing Technology, Shenzhen University, Shenzhen, China.
His main research involves computer vision and machine learning, with a long-term focus on image anomaly detection, image generation, and fast retrieval.
\end{IEEEbiography}

\begin{IEEEbiography}[{\includegraphics[width=1in,height=1.25in,clip,keepaspectratio]{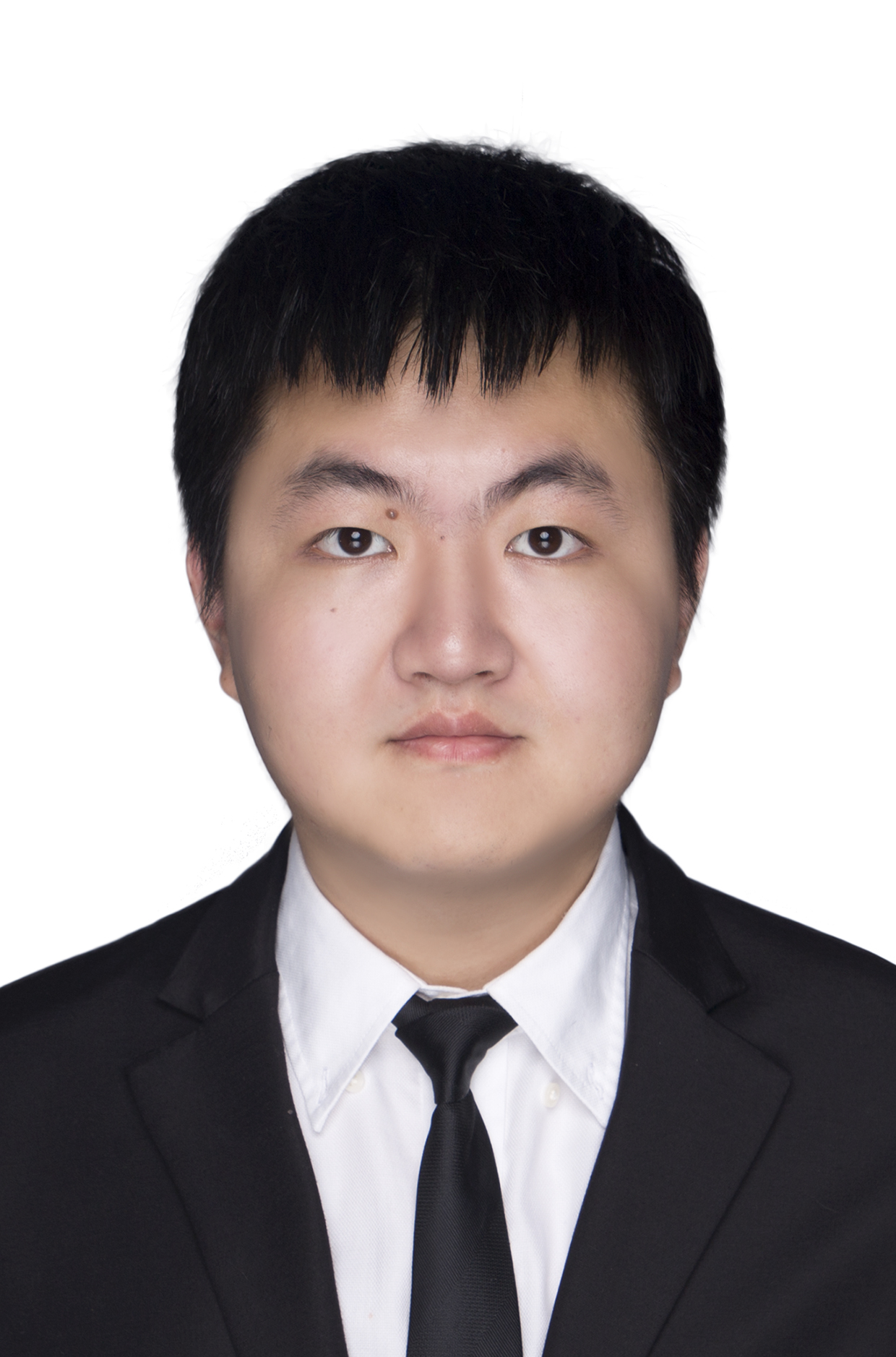}}]{Jiaqi Liu} received his BS degree from Dalian University of Technology in 2019. He is pursuing the MS degree from Southern University of Science and Technology under the supervision of Professor Feng Zheng. His research interest focuses on image anomaly detection.
\end{IEEEbiography}

\begin{IEEEbiography}[{\includegraphics[width=1in,height=1.25in,clip,keepaspectratio]{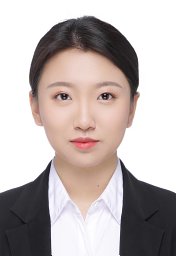}}]{Jiayi Lyu} was born in Ningxia, China, in 1999. She graduated from Capital Normal University, Beijing, China, and received the bachelor’s degree in July 2021. She is currently pursuing the Ph.D. degree in computer applied technology with the School of Engineering Science, University of Chinese Academy of Sciences, Beijing. Her research interests include image processing and computer graphics.
\end{IEEEbiography}

\begin{IEEEbiography}[{\includegraphics[width=1in,height=1.25in,clip,keepaspectratio]{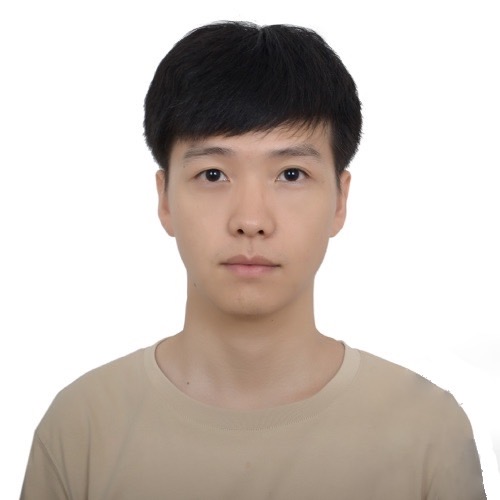}}]{Yong Liu} received the B.S. degree in software engineering from North China Electric Power University, Beijing, China, in 2015, and the M.S. degrees in computer science from the Institute of Computing Technology (ICT), Chinese Academy of Science (CAS), Beijing, China, in 2018. His interests include computer vision, pattern recognition, and machine learning.
\end{IEEEbiography}

\begin{IEEEbiography}[{\includegraphics[width=1in,height=1.25in,clip,keepaspectratio]{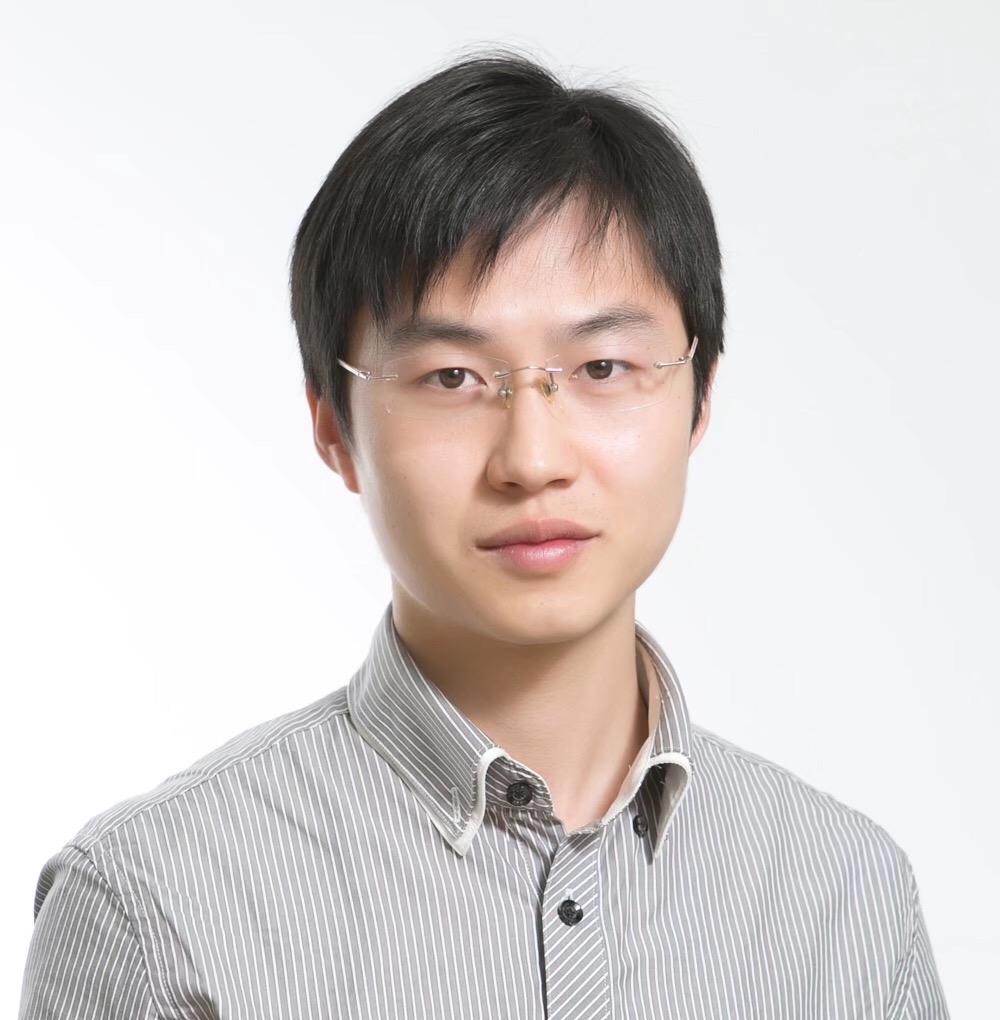}}]{Chengjie Wang} received the B.S. degree in computer science from Shanghai Jiao Tong University, China, in 2011, and double M.S. degrees in computer science from Shanghai Jiao Tong University, China and Weseda University, Japan, in 2014. He is currently the Research Director of Tencent YouTu Lab. His research interests include computer vision and machine learning. He has published more than 100 papers on major Computer Vision and Artificial Intelligence Conferences such as CVPR, ICCV, ECCV, AAAI, IJCAI and NeurIPS, and holds over 100 patents in these areas.
\end{IEEEbiography}

\begin{IEEEbiography}[{\includegraphics[width=1in,height=1.25in,clip,keepaspectratio]{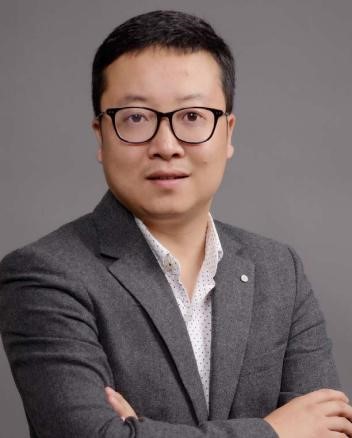}}]{Feng Zheng} received the Ph.D. degree from The University of Sheffield, Sheffield, U.K., in 2017. He is currently an Assistant Professor with the Department of Computer Science and Engineering, Southern University of Science and Technology, Shenzhen, China. His research interests include machine learning, computer vision, and human-computer interaction.
\end{IEEEbiography}

\begin{IEEEbiography}[{\includegraphics[width=1in,height=1.25in,clip,keepaspectratio]{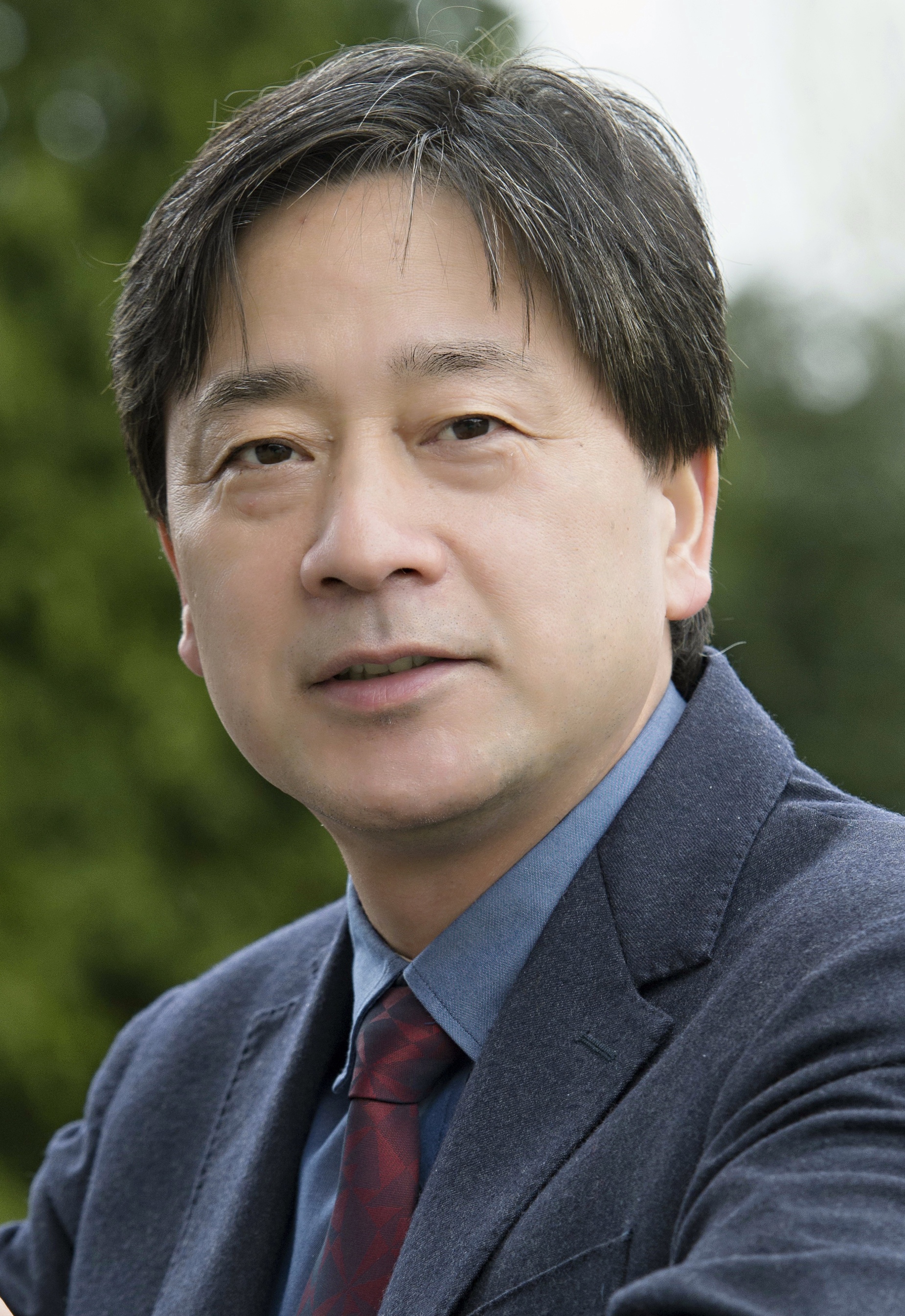}}]{Yaochu Jin} (Fellow, IEEE) received the B.Sc., M.Sc., and Ph.D. degrees from Zhejiang University, Hangzhou, China, in 1988, 1991, and 1996, respectively, and the Dr.-Ing. degree from Ruhr University Bochum, Germany, in 2001. He is presently Chair Professor of AI, Head of the Trustworthy and General AI Lab, School of Engineering, Westlake University, Hangzhou, China. Prior to that, he was Alexander von Humboldt Professor for Artificial Intelligence endowed by the German Federal Ministry of Education and Research, with the Faculty of Technology, Bielefeld University, Germany from 2021 to 2023, and Surrey Distinguished Chair, Professor in Computational Intelligence, Department of Computer Science, University of Surrey, Guildford, U.K. from 2010 to 2021.  He was also “Finland Distinguished Professor” with University of Jyväskylä, Finland, and “Changjiang Distinguished Visiting Professor” with the Northeastern University, China from 2015 to 2017. His main research interests include evolutionary optimization, evolutionary machine learning, trustworthy AI, and evolutionary developmental AI. Prof Jin is the President of the IEEE Computational Intelligence Society and Editor-in-Chief of Complex \& Intelligent Systems. He was named by the Web of Science as “a Highly Cited Researcher” from 2019 to 2023 consecutively. He is a Member of Academia Europaea.

\end{IEEEbiography}



\end{document}